\documentclass[10pt,twocolumn,letterpaper]{article}

\usepackage{cvpr}              
\definecolor{cvprblue}{rgb}{0.21,0.49,0.74}
\definecolor{HeaderBlue}{RGB}{234,242,255}
\usepackage[pagebackref,breaklinks,colorlinks,allcolors=cvprblue]{hyperref}

\usepackage{microtype}
\usepackage{graphicx}
\usepackage{booktabs} 
\usepackage{multirow}
\usepackage{pifont}
\usepackage{stmaryrd}
\usepackage{mathtools}

\usepackage[hypcap=false]{caption}
\usepackage{graphicx}
\usepackage{cuted}
\usepackage{capt-of}
\usepackage{makecell}

\usepackage{colortbl}
\usepackage[table]{xcolor}
\usepackage{booktabs}

\definecolor{brickred}{rgb}{0.8, 0.25, 0.33}
\definecolor{brickgreen}{rgb}{0.25, 0.8, 0.33}
\newcommand{\cm}{\textcolor{brickgreen}{\ding{51}}}%
\newcommand{\xm}{\textcolor{brickred}{\ding{55}}}%
\usepackage[table,dvipsnames]{xcolor}         
\usepackage{color}
\definecolor{brickred}{rgb}{0.8, 0.25, 0.33}
\definecolor{brickred2}{rgb}{0.25, 0.8, 0.33}
\usepackage{graphicx}

\usepackage{etoolbox}

\newcommand{\ttabref}[1]{Tab.~\ref{#1}}
\newcommand{\ffigref}[1]{Fig.~\ref{#1}}
\newcommand{\ssecref}[1]{Sec.~\ref{#1}}
\newcommand{\eeqref}[1]{Eq.~(\ref{#1})}

\newcommand{\framework}{PNG}

\def\paperID{} 
\def\confName{CVPR}
\def\confYear{2026}

\title{Diffusion-Based sRGB Real Noise Generation \\ via Prompt-Driven Noise Representation Learning}

\author{
Jaekyun Ko$^{1,2}$\thanks{Equal contribution.} \quad
Dongjin Kim$^{1}$\footnotemark[1] \quad
Soomin Lee$^{1}$ \quad
Guanghui Wang$^{3}$ \quad
Tae Hyun Kim$^{1}$\thanks{Corresponding author.} \\
$^{1}$Department of Computer Science, Hanyang University \\
$^{2}$Mobile Experience (MX) Division, Samsung Electronics \\
$^{3}$Department of Computer Science, Toronto Metropolitan University \\
{\tt\small jkko1124.ko@samsung.com 
\{dongjinkim,min001017,taehyunkim\}@hanyang.ac.kr wangcs@torontomu.ca} \\
\tt\small \href{https://github.com/JK-the-Ko/PNG}{https://github.com/JK-the-Ko/PNG}
}

\begin{document}
\maketitle

\begin{strip}
  \centering
  \begin{minipage}{\textwidth}
    \centering
    \vspace{-18mm}
    \includegraphics[width=1.0\textwidth]{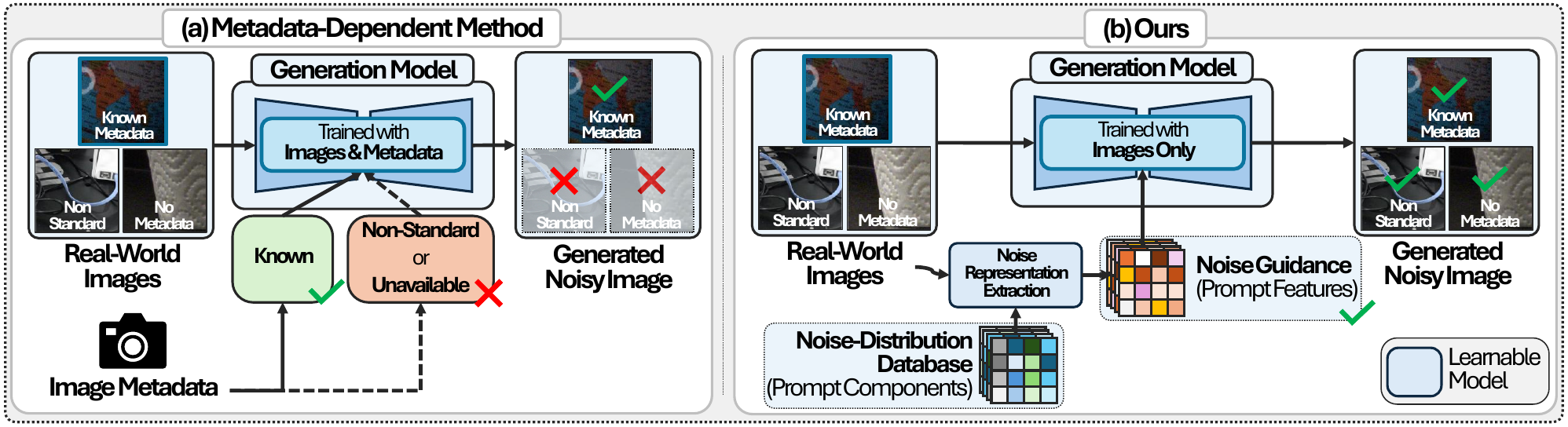}
    \vspace{-7.0mm}
    \captionof{figure}{Noise Generation Comparison. (a) Metadata-dependent approach. (b) Ours with metadata-free approach.}
    \label{fig:teaser}
    \vspace{-3.5mm}
  \end{minipage}
\end{strip}

\begin{abstract}
Denoising in the sRGB image space is challenging due to large noise variability.
Although end-to-end methods perform well, their effectiveness in real-world scenarios is limited by the scarcity of real noisy-clean image pairs, which are expensive and difficult to collect. To address this limitation, several generative methods have been developed to synthesize realistic noisy images from limited data. 
These approaches often rely on camera metadata during both training and testing to synthesize real-world noise. 
However, the lack of metadata or inconsistencies between devices restricts their usability.
Therefore, we propose a novel framework called \textbf{P}rompt-Driven  \textbf{N}oise \textbf{G}eneration (\textbf{\framework{}}). This model is capable of acquiring high-dimensional prompt features that capture the characteristics of real-world input noise and creating a variety of realistic noisy images consistent with the distribution of the input noise.
By eliminating the dependency on explicit camera metadata, our approach significantly enhances the generalizability and applicability of noise synthesis.
Comprehensive experiments reveal that our model effectively produces realistic noisy images and show the successful application of these generated images in removing real-world noise across various benchmark datasets.
\end{abstract}

\vspace{-12.5mm}

\section{Introduction}
\label{introduction}
Real-world denoising is particularly challenging in low-level vision tasks due to the inherent complexity and diversity of real noise.
Unlike additive white Gaussian noise (AWGN), which can be effectively modeled under controlled laboratory conditions, real-world noise arises from various sources such as sensor imperfections, lighting variations, in-camera processing pipelines, and adjustable camera settings. Consequently, real-world noise frequently exhibits signal-dependent and spatially varying properties.

A common solution to this issue is to construct extensive paired datasets consisting of noisy-clean image pairs and subsequently train end-to-end denoising networks in a supervised manner~\cite{mirnet, hinet, restormer, nafnet, idf}. However, collecting such datasets is both resource-intensive and technically demanding~\cite{SIDD, siddplus, polyu, nam, dnd}. 
To avoid this bottleneck, recent approaches generate synthetic real-world noise using metadata which includes sensor information such as camera manufacturer, ISO settings, shutter speed, and other parameters~\cite{noiseflow, Flow-sRGB, cycleisp, naflow, interflow, degflow}. This metadata serves as a compact representation of how an image signal processor (ISP) transforms a RAW image to an sRGB one, thereby enabling an accurate simulation of the noise characteristics.

Despite their effectiveness, metadata-dependent approaches~\citep{noiseflow, Flow-sRGB, naflow, neca} encounter practical limitations. As illustrated in~\ffigref{fig:teaser} (a), publicly available sRGB images on websites often lack original EXIF tags due to post-processing, resulting in metadata being unavailable. In domains like scientific imaging, metadata semantics and formats vary greatly or may be absent, making standard metadata methods ineffective. Therefore, approaches relying on specific metadata formats struggle to generalize well across diverse real-world scenarios.

Hence, in this work, we propose \textbf{P}rompt-Driven  \textbf{N}oise \textbf{G}eneration (\textbf{\framework{}}),
a novel framework that eliminates the dependency on explicit metadata during both training and noise generation stages.
As illustrated in \ffigref{fig:teaser}(b), we propose an implicit noise representation extraction utilizing prompt features, 
capturing input-specific noise characteristics directly from input noise images. These prompt features are generated via learnable prompt components, 
serving as a comprehensive repository of noise distributions. 
By conditioning the model on these learned prompt features, our framework enables a unified noise generator capable of synthesizing realistic noise. This design alleviates the dependency on metadata, which is often inconsistent, non-standard, or unavailable in both training and inference phases.

Specifically, we introduce a \textbf{P}rompt \textbf{A}uto\textbf{E}ncoder (\textbf{PAE}) that encodes the noise of the input image and produces input-specific prompt features including characteristics of the input noise, such as ISO levels (gain) and noise correlation patterns.
We then use a consistency model (CM)~\citep{cm, latentcm}, a diffusion-based generative model, to learn the latent space of the PAE and synthesize a new latent code conditioned on the extracted high-dimensional prompt features, which encapsulate information about the noise characteristics.
Precisely, building on recent diffusion transformer (DiT)~\citep{dit} architectures based on CM, we present a \textbf{P}rompt \textbf{DiT} (\textbf{P-DiT}) that fully leverages the prompt features from the PAE during the generation process.
Finally, the Decoder of our PAE takes the synthesized latent code from P-DiT together with the clean image, and produces a noisy image, thereby simulating signal-dependent real-world noise.
Through extensive experiments, we demonstrate that our \framework{} achieves outstanding real-world noise modeling quality without requiring any metadata during the training or testing phases.
For downstream tasks, we trained a conventional supervised image denoising network on the generated noisy dataset, achieving state-of-the-art (SOTA) real-world denoising performance across diverse domains, including smartphones, and DSLR cameras.



\section{Related Work}
\label{related_works}

\noindent\textbf{sRGB Noise Generation.}
The limited availability of real-world noise datasets results in overfitting issues for denoising networks, particularly in practical applications~\citep{noiseflow,cycleisp,c2n,noisediffusion,apbsn, mmbsn, lgbpn, maskedshuffle, iterative_denoising, randomsubsample}. 
To tackle this problem, various methods for generating sRGB noise have been proposed.
Flow-sRGB~\citep{Flow-sRGB} leverages normalizing flows~\cite{variational_nf, realnvp,glow} to model noise distributions based on factors like smartphone type and gain settings.
NeCA~\citep{neca} introduces a neighboring correlation-aware noise model for synthesizing realistic noise, explicitly accounting for both signal dependency and neighboring noise correlations.
However, these methods depend on explicit metadata for noise generation such as smartphone device, limiting their applicability in situations where such information is unavailable.
To address this, NAFlow~\citep{naflow} introduces a noise-aware sampling algorithm to synthesize sRGB noise without requiring metadata during inference. However, it still requires metadata during the training phase. 
In contrast to these previous approaches, our approach offers a metadata-free noise generation method for both training and testing phases by leveraging prompt features as noise guidance that capture input-noise-specific information.

\noindent\textbf{Prompt Learning.}
In natural language processing, prompts (brief textual snippets) are prepended or appended to input sequences to guide large language models (LLMs) toward specific behaviors or outputs~\citep{openai_gpt3,autoprompt,pet}. These prompts offer an auxiliary context, effectively transforming downstream tasks into instances that closely resemble the original pre-training objectives of the model.

Recently, this concept has been extended into computer vision by substituting fixed textual prompts with learnable prompts, either embedded into the input sequences or intermediate layers.
The lightweight prompt parameters allow models to efficiently adapt to new domains or tasks~\citep{pip_universal, promptir, promptrestorer, context_optimization, ucip, promptcir, promptrr, segprompt, freprompter}.
For example, PromptIR~\citep{promptir}, an all-in-one low-level image restoration framework, introduces a compact set of learnable prompts that dynamically interact with degradation features, providing specific degradation guidance and efficiently adapting the model to various restoration tasks.

Inspired by these works, we conceptualize prompts as compact representations of real-world noise characteristics. Our proposed PAE extracts high-dimensional prompt features that capture detailed sensor-specific statistics such as ISO levels, spatial correlations, and other noise attributes. Subsequently, our P-DiT leverages these prompt features to conditionally generate realistic sRGB noise. Thus, our method replaces explicit metadata with prompt-based representations, enabling precise control and substantially improving both the realism and diversity of the synthesized noisy images.


\begin{figure*}[t]
\begin{center}
\centerline{\includegraphics[width=2.45\columnwidth]{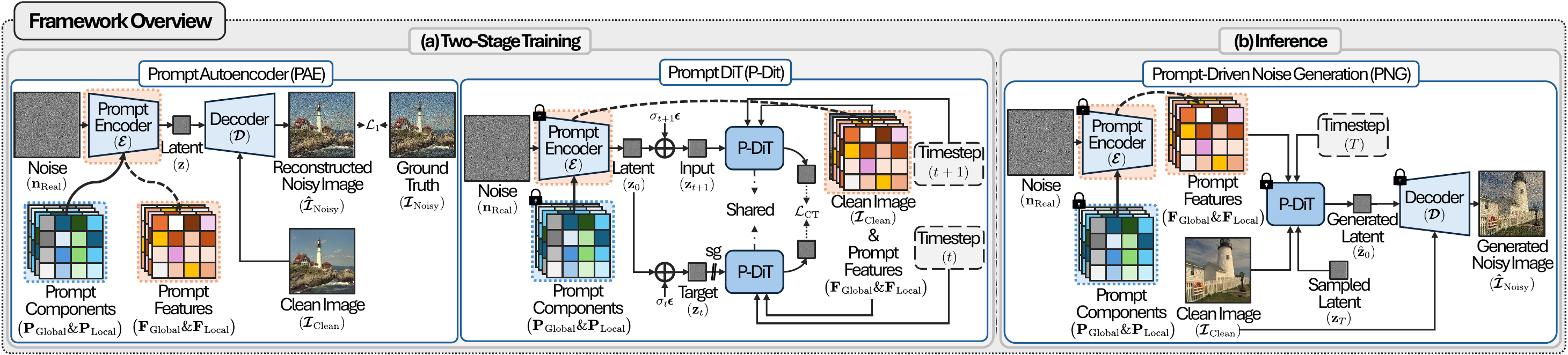}}
\caption{Overview of the proposed method. (a) Training pipeline. (b) Inference pipeline.}
\label{fig:overall_flow}
\vspace*{-12mm}
\end{center}
\end{figure*}

\section{Proposed Method}
\subsection{Preliminaries}
\label{sec:3.1}
\noindent\textbf{Diffusion Models.}
Diffusion models~\citep{sohdiffusion, songscore1, hodiffusion, diffir, diffbir, seesr, onestepsr} involve a forward process that corrupts the data with independent and identically distributed (\textit{i.i.d.}) Gaussian noise, followed by a learned reverse process.
In the forward process, random Gaussian noise is added to the given input $\mathbf{x}_0$.
Using reparameterization, this process at timestep $t$ can be represented as:
\begin{equation} 
\mathbf{x}_t = \alpha_t \mathbf{x}_0 + \sigma_t \pmb{\epsilon}, \quad \pmb{\epsilon} \sim \mathcal{N}(0, \mathrm{\mathbf{I}}), \quad t  \in \{0, 1, \dots, T\},
\label{eq:1}
\end{equation} 
where $\mathbf{x}_0$ denotes the clean sample and $\mathbf{x}_t$ is the corresponding noisy sample at timestep $t$. 
Here, $\alpha_t$ controls the proportion of the original signal preserved, 
while $\sigma_t$ defines the noise schedule that determines the noise magnitude. 
The noise $\pmb{\epsilon}$ is drawn from a standard normal distribution.
Different parameterizations of $\alpha_t$ and $\sigma_t$ define distinct diffusion processes, including variance-preserving~\citep{sohdiffusion, hodiffusion} and variance-exploding (VE)~\citep{songscore1, songscore3, vdm} processes.
In this work, we employ the VE diffusion process (\ie, $\alpha_t\text{=}1$).
The objective of diffusion models is to reverse the forward process by denoising the corrupted sample $\mathbf{x}_t$ back to the original data $\mathbf{x}_0$.

In the context of score-based generative models~\citep{songscore1,songscore2,songscore3}, the score model $s_\theta(\mathbf{x}_{t}, \sigma_t)$ estimates the score function, a vector field that points toward regions of higher data density, by employing score matching~\citep{scorematching,songscore3}.
The reverse process is performed by iteratively applying the learned score function to gradually denoise the noisy input.
This can be represented using probability flow ordinary differential equations (PF ODEs), as described in~\citep{edm} as:
\begin{equation}
d\mathbf{x}_t = -\dot \sigma_{t} \sigma_{t} \nabla_{\mathbf{x}_{t}} \log p(\mathbf{x}_{t}; \sigma_{t})dt,
\end{equation} 
where $\dot{\sigma}_t$ denotes the time derivative of $\sigma_t$, and the score function is $\nabla_{\mathbf{x}_{t}} \log p(\mathbf{x}_{t}; \sigma_{t})$. 
In practice, solving this reverse-time ODE involves numerous update steps, which makes diffusion models computationally intensive.

\noindent\textbf{Consistency Models.}
Unlike traditional diffusion models that generate data through multiple iterative steps, recent consistency models (CMs) streamline the process by achieving results in a single step~\citep{cm, ict}.
Specifically, the CM learns a mapping function $f_\theta(\mathbf{x}_t, \sigma_t)$ such that, for any $\mathbf{x}_t$, the output remains \textit{consistent} with the original data $\mathbf{x}_0$. This process can be approximated as follows:
\begin{equation}
f_\theta(\mathbf{x}_t, \sigma_t) = \mathbf{x}_t + \int_{\sigma_t}^{\sigma_0} \frac{d\mathbf{x}_u}{du} \,du \approx \mathbf{x}_0.
\end{equation}
The consistency training (CT) loss ensures that the model's outputs remain stable across different noise levels, as follows:
\begin{equation}
\mathcal{L}_\text{CT} = \mathbb{E} \left[ \lambda(\sigma_t) \, d \left( f_\theta(\mathbf{x}_{t+1}, \sigma_{t+1}), \text{sg}(f_{\theta^{-}}(\mathbf{x}_{t}, \sigma_{t})) \right) \right],
\label{eq:ct_loss}
\end{equation}
where $\lambda(\cdot)$ is a weighting function, $d(\cdot)$ is a distance function like pseudo-Huber loss~\citep{ict}, and $\text{sg}(\cdot)$ denotes the stop-gradient operator.
Here, $f_\theta$ and $f_{\theta^{-}}$ refer to the student and teacher networks, respectively.
The stop-gradient operator is used to keep the teacher network fixed during each optimization step of the student network.
In our approach, since the pretrained teacher model is unavailable, we set $\theta^{-}=\theta$ as in~\citep{cm, ict, actdiffusion}.
We adopt most of the training hyperparameters for the CM from EDM~\citep{edm} and iCT~\citep{ict}, and further details are in \ssecref{appendix:training_details}.

\subsection{Overall Flow: PNG}
\label{sec:3.2}
    We introduce a novel Prompt-Driven Noise Generation framework (\framework{}), designed to generate realistic noisy images by substituting explicit metadata (\eg, smartphone manufacturer, ISO settings) with learned prompt components. Inspired by recent latent diffusion models~\citep{ldm, latentcm}, \framework{} operates within a compact latent space, achieving significant computational efficiency via CM. This approach notably reduces computational overhead compared to methods operating directly in image space. The overall flow of \framework{} is depicted in \ffigref{fig:overall_flow}, comprising two primary components: the Prompt Autoencoder (PAE) and the Prompt DiT (P-DiT), trained sequentially in a two-stage process.

\begin{figure*}[t]
\begin{center}
\centerline{\includegraphics[width=2.125\columnwidth]{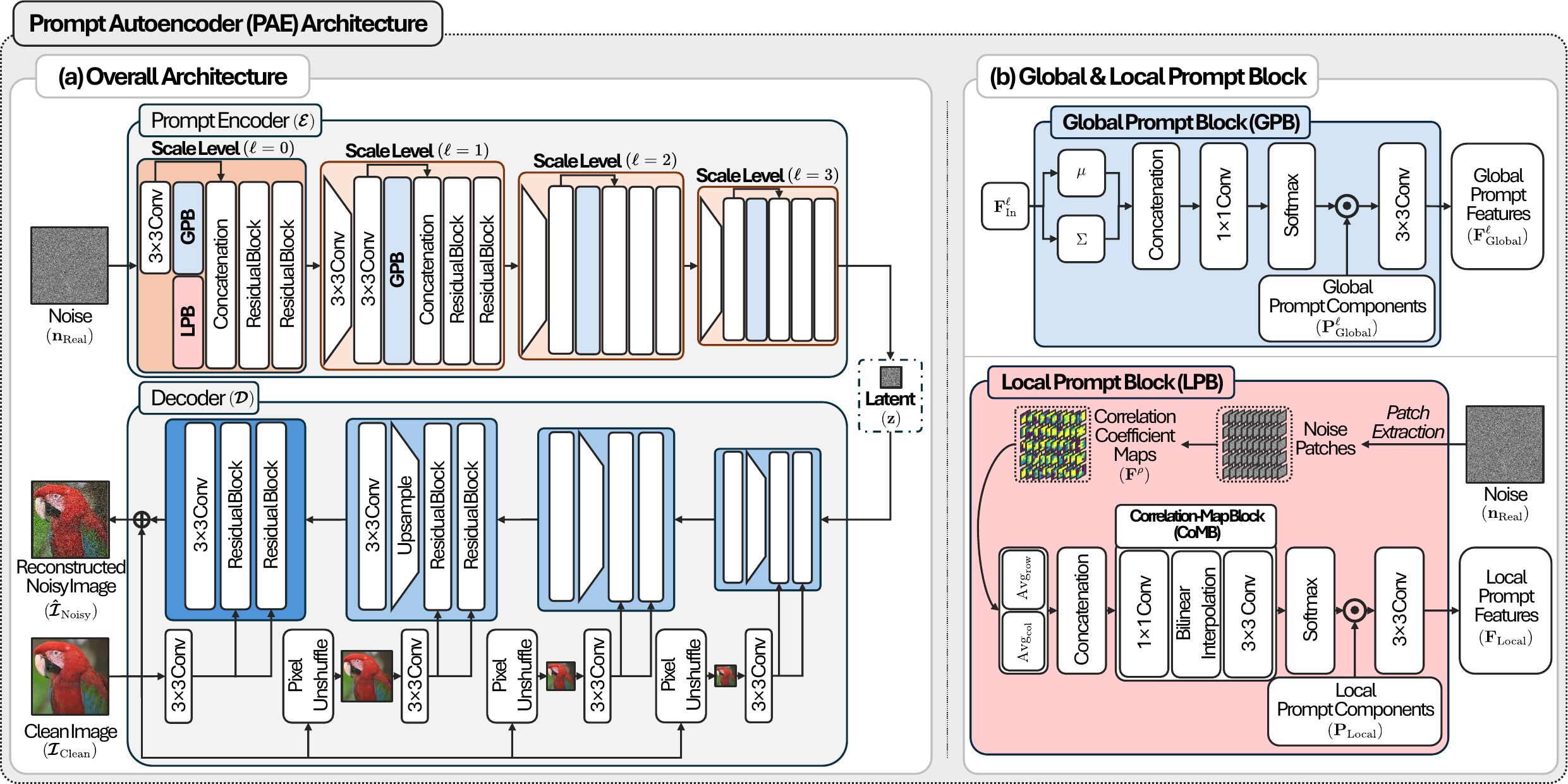}}
\caption{(a) Sketch of the Prompt Autoencoder ({PAE}). (b) Details of Global and Local Prompt Blocks.}
\label{fig:arch_pe}
\vspace*{-12mm}
\end{center}
\end{figure*}

\noindent\textbf{Two-Stage Training Phase.}
    First, we train a U-Net-based~\citep{unet} PAE to learn a compact latent representation. As illustrated in \ffigref{fig:overall_flow} (a), the PAE consists of a Prompt Encoder $\mathcal{E}$ and a Decoder $\mathcal{D}$. 
    
    The Prompt Encoder $\mathcal{E}$ maps the input noise \text{\footnotesize $\mathrm{\mathbf{n}}_\mathrm{Real}$}, representing the residual between the paired noisy image \text{\footnotesize $\pmb{\mathcal{I}}_\mathrm{Noisy}$} and the clean image \text{\footnotesize $\pmb{\mathcal{I}}_\mathrm{Clean}$} (\ie, \text{\footnotesize $\mathrm{\mathbf{n}}_\mathrm{Real} = \pmb{\mathcal{I}}_\mathrm{Noisy} - \pmb{\mathcal{I}}_\mathrm{Clean}$}), into a latent code \text{\footnotesize $\mathrm{\mathbf{z}}$}. 
    Moreover, it learns real-world noise characteristics from the training dataset through prompt components $\mathbf{P}_{\mathrm{Global}}$ and $\mathbf{P}_{\mathrm{Local}}$ which are learnable parameters jointly optimized with the PAE. 
    The PAE then generates prompt features $\mathbf{F}_{\mathrm{Global}}$ and $\mathbf{F}_{\mathrm{Local}}$ that capture the characteristics of the input noise $\mathbf{n}_{\mathrm{Real}}$ by modulating the prompt components conditioned on the input noise.
    Given the clean image \text{\footnotesize $\pmb{\mathcal{I}}_\mathrm{Clean}$}, the Decoder $\mathcal{D}$ reconstructs the noisy image $\hat{\pmb{\mathcal{I}}}_\mathrm{Noisy}$ from the latent code $\mathbf{z}$, learning signal-dependent characteristics simultaneously.
    
    Subsequently, given the fully trained Prompt Encoder $\mathcal{E}$, the P-DiT is trained to learn the latent distribution of the PAE using the CM-based objective $\mathcal{L}_{\text{CT}}$ defined in~\eeqref{eq:ct_loss}, conditioned on the clean image ${\pmb{\mathcal{I}}}_{\mathrm{Clean}}$ and the prompt features.

\textbf{Inference Phase.}
    As illustrated in \ffigref{fig:overall_flow} (b), during the generation phase, the Prompt Encoder $\mathcal{E}$ produces prompt features that capture the characteristics of input noise $\mathrm{\mathbf{n}}_\mathrm{Real}$.
    Given the resulting prompt features and a clean image, P-DiT transforms a Gaussian noise sample $\mathbf{z}_T$ into a latent code $\hat{\mathbf{z}}_{0}$. 
    The Decoder $\mathcal{D}$ then converts $\hat{\mathbf{z}}_{0}$ into the final noisy image $\hat{\pmb{\mathcal{I}}}_\mathrm{Noisy}$, conditioned on the clean image ${\pmb{\mathcal{I}}}_\mathrm{Clean}$.

\vspace{-1.5mm}

\subsection{Prompt Autoencoder}
\label{sec:3.3}
\subsubsection{Prompt Encoder}
    As depicted in \ffigref{fig:arch_pe}, the Prompt Encoder $\mathcal{E}$ is composed of several convolutional layers, including residual blocks~\citep{resnet}, and incorporates two key modules: the Global Prompt Block (GPB) and the Local Prompt Block (LPB). These modules dynamically generate prompt features by leveraging learnable prompt components guided by extracted real-world noise characteristics, such as ISO levels and noise correlation. The resulting prompt features are concatenated with the input features, embedding informative latent codes. Through training, the learnable prompt components within both GPB and LPB adaptively capture the underlying data distribution, allowing the prompt features to serve as implicit representations of input noise characteristics, thereby effectively replacing the need for explicit noise metadata.

\noindent\textbf{Global Prompt Block.}
A camera ISP can adjust brightness and sensitivity using ISO (gain), simulating physical exposure. However, increasing the signal gain through high ISO settings also amplifies sensor noise. 
Thus, ISO is a crucial factor in modeling noise attributes.
To capture global noise statistics, such as noise amplification in the prompt features, we propose the Global Prompt Block (GPB).
This block is designed to analyze how different ISO levels impact noise characteristics, allowing our model to better understand the inherent noise patterns associated with varying ISO settings.

In the GPB, we first define a set of learnable global prompt components \text{\footnotesize $\mathrm{\mathbf{P}}^\ell_{\mathrm{Global}}\in\mathbb{R}^{{\frac{H}{2^\ell}}{\times}{\frac{W}{2^\ell}}{\times}C^\ell_\mathrm{Global}}$} to capture the global statistics at scale level $\ell\in\{0,1,2,3\}$, where $H$, $W$, and $C$ represent the height, width, and channel dimension of the corresponding feature maps, respectively.
To generate the global prompt features \text{\footnotesize $\mathbf{F}^\ell_\mathrm{Global}\in\mathbb{R}^{{\frac{H}{2^\ell}}{\times}{\frac{W}{2^\ell}}{\times}C^\ell_\mathrm{Global}}$}, we compute input-specific coefficients \text{\footnotesize $\mathbf{w}^\ell_{\mathrm{Global}}\in\mathbb{R}^{1{\times}1{\times}C^\ell_\mathrm{Global}}$} from the input feature \text{\footnotesize $\mathbf{F}^\ell_\mathrm{In}\in\mathbb{R}^{{\frac{H}{2^\ell}}{\times}{\frac{W}{2^\ell}}{\times}C^\ell_\mathrm{In}}$} by computing the channel-wise mean and standard deviation of the input feature to capture the global statistics of the input noise as:
\begin{equation}
\mathbf{w}^\ell_{\mathrm{Global}}=\mathrm{Softmax}\bigg(\mathrm{Conv}_{1\times1}\big\llbracket\mu(\mathbf{F}^\ell_\mathrm{In}),\Sigma(\mathbf{F}^\ell_\mathrm{In})\big\rrbracket\bigg),
\end{equation}
where $\mu(\cdot)$ and $\Sigma(\cdot)$ represent functions that compute the channel-wise mean and standard deviation, respectively. These two results are concatenated and passed through a $1{\times}1$ convolutional layer, followed by a softmax operation.
Then, these coefficients are adopted to dynamically modify the global prompt components \text{\footnotesize $\mathrm{\mathbf{P}}^\ell_{\mathrm{Global}}$}, followed by a 3${\times}$3 convolutional layer for refinement, to yield the final global prompt features \text{\footnotesize $\mathbf{F}^\ell_\mathrm{Global}$} as:
\begin{equation}
\mathrm{\mathbf{F}}^\ell_{\mathrm{Global}}=\mathrm{Conv}_{3\times3}\bigg(\mathbf{w}^\ell_{\mathrm{Global}}\odot\mathrm{\mathbf{P}}^\ell_{\mathrm{Global}}\bigg),
\end{equation}
where $\odot$ denotes the element-wise multiplication.
This process ensures that the global prompt features reflect the global noise characteristics based on noise distribution statistics in the input image.

\noindent\textbf{Local Prompt Block.}
In the GPB, we extract global information (\eg, ISO); however, real-world sRGB noise cannot be fully characterized by global information alone.
This issue arises from several transformations in the ISP pipeline, such as non-linear and spatially varying operations, 
which cause patterned and non-\textit{i.i.d.}
noise during the RAW-to-sRGB conversion.
Therefore, we additionally propose the Local Prompt Block (LPB) to capture camera model-specific noise characteristics from the input noise.

In the LPB, we first extract a $\rho{\times}\rho$ patch from the input noise \text{\footnotesize $\mathrm{\mathbf{n}}_\mathrm{Real}\in\mathbb{R}^{H{\times}W{\times}3}$} at every pixel location and calculate the local correlation map for each patch with respect to its center pixel. 
Specifically, following the approach in LGBPN~\citep{lgbpn}, we compute the Pearson correlation coefficients of neighboring pixels relative to the center pixel for each patch.
This operation is applied to all patches, resulting in correlation coefficient maps \text{\footnotesize $\mathrm{\mathbf{F}}^\rho\in\mathbb{R}^{H{\times}W{\times}\rho^2}$}, where each pixel contains its local correlation information along the channel axis. 
We then separately compute row-wise and column-wise averages from \text{\footnotesize $\mathrm{\mathbf{F}}^\rho$} 
to extract spatial correlations of real-world noise introduced by the ISP pipeline~\cite{apbsn, lgbpn, noise2void}.
These two averages are then concatenated and upscaled using correlation map block (CoMB) which consists of lightweight operations, including a 1${\times}$1 convolutional layer, a bilinear upsampler, and a 3${\times}$3 convolutional layer, as illustrated in~\ffigref{fig:arch_pe} (b).
Similar to the GPB, we apply a softmax operation to the upscaled features to generate the local prompt coefficients 
\text{\footnotesize$\mathbf{w}_{\mathrm{Local}}\in\mathbb{R}^{H{\times}W{\times}C_\mathrm{Local}}$} defined as
\begin{equation}
\footnotesize
\mathbf{w}_{\mathrm{Local}}=\mathrm{Softmax}\bigg(\mathrm{CoMB}\big(\big\llbracket\mathrm{\mathrm{Avg}_{\mathrm{row}}(\mathbf{F}}^\rho),\mathrm{\mathrm{Avg}_{\mathrm{col}}(\mathbf{F}}^\rho)\big\rrbracket\big)\bigg),
\end{equation}
where \text{\footnotesize $\mathrm{Avg}_{\mathrm{row}}$} and \text{\footnotesize $\mathrm{Avg}_{\mathrm{col}}$} indicate row-wise and column-wise averaging operations, respectively.
Next, the local prompt coefficients are multiplied element-wise by the prompt components \text{\footnotesize $\mathrm{\mathbf{P}}_{\mathrm{Local}}\in\mathbb{R}^{H{\times}W{\times}C_\mathrm{Local}}$} for dynamic feature aggregation, yielding the local prompt features \text{\footnotesize $\mathrm{\mathbf{F}}_\mathrm{Local}\in\mathbb{R}^{H{\times}W{\times}C_\mathrm{Local}}$} as follows:
\begin{equation}
\mathrm{\mathbf{F}}_{\mathrm{Local}}=\mathrm{Conv}_{3\times3}\bigg(\mathbf{w}_{\mathrm{Local}}\odot\mathrm{\mathbf{P}}_\mathrm{Local}\bigg).
\end{equation}
We emphasize that by leveraging local noise characteristics (\eg, noise local correlation), the prompt components learn to focus on unique 
camera-specific features.

\subsubsection{Decoder}
As illustrated in the lower part of \ffigref{fig:arch_pe} (a), the Decoder $\mathcal{D}$ consists of multiple convolutional layers, integrating residual blocks and upsampling operations to transform latent codes back into noisy images. 
The upsampling operation increases the spatial resolution of input features at scale $\ell$ using nearest-neighbor upsampling, following~\citep{stablediffusion}. 
To further capture the signal-dependent characteristics of real-world noise, clean images are downsampled using pixel downsampling~\citep{pixelshuffle}, preserving fine-grained textures and conditioning at each scale level.

\subsection{Prompt DiT (P-DiT)}
We introduce P-DiT, which fully leverages the prompt features extracted from the Prompt Encoder $\mathcal{E}$ to synthesize latent codes $\mathbf{\hat{z}}_0$ that align with the embedded information of the input noise characteristics and the clean image provided.
Our P-DiT is based on DiT~\citep{dit}, a transformer-based CM architecture specifically designed for training diffusion models. See~\ssecref{sec_pdit} in the supplementary material for detailed P-DiT architecture description.

\vspace{-2.5mm}

\section{Experiments}
\subsection{Experimental Setup}
\label{sec:4.1}
\noindent\textbf{Implementation Details}
The PAE is trained with the Adam optimizer~\citep{adam} and minimizes the $\mathcal{L}_{1}$ loss between the noisy image \text{\footnotesize $\pmb{\mathcal{I}}_\mathrm{Noisy}$} and the reconstructed noisy image \text{\footnotesize $\hat{\pmb{\mathcal{I}}}_\mathrm{Noisy}$}, along with $\mathcal{L}_{2}$ regularization applied to the latent code $\mathrm{\mathbf{z}}$ to encourage a dense representation~\citep{stablediffusion}.
We start with an initial learning rate of 1e\text{-}4, which is then reduced to 1e\text{-}6 using a cosine annealing algorithm~\citep{cosine_annealing} over 400k iterations.
We use randomly cropped patches of size $256{\times}256$ and a mini-batch size of 64 for training. 

The P-DiT model is optimized with the RAdam optimizer~\citep{radam} using a fixed learning rate of 2e\text{-}4 over 250k iterations, applying the pseudo-Huber loss for \eeqref{eq:ct_loss}, similar to iCT~\citep{ict}. We also use randomly cropped patches of size $256{\times}256$ to embed latent codes of size $ 32{\times}32 $, with a mini-batch size of 512 during training.
For the ablations, the mini-batch size is reduced to 128 to lower training cost.
More details on the training process and model configurations are in \ssecref{appendix:training_details} and \ssecref{appendix:model_capacity}, respectively.

For real-world sRGB image denoising, we employ the DnCNN architecture \citep{dncnn}, trained for 100k iterations with the Adam optimizer. All hyperparameters follow the setup in~\citep{neca,naflow}, using a learning rate of 1e\text{-}3, patch size of $96\times96$, and a mini-batch size of 8. 

\begin{table}[!t]
    \centering
\caption{Quantitative results for synthetic noise on a subset of the SIDD validation set, in which the ISO values exist in the training set. The results are computed with KLD$\downarrow$ and AKLD$\downarrow$. The best and second-best results are highlighted in \textbf{bold} and \underline{underline}, respectively.}
    \vspace{-2.5mm}
    \resizebox{1.0\columnwidth}{!}{
    \begin{tabular}{cl|ccccc}
    \toprule[0.5pt]
     
    Camera                   & Metrics & C2N    & Flow-sRGB & NeCA-W & NAFlow & \textbf{Ours} \\
    \midrule[0.2pt]
    \multirow{2}{*}{G4}      & KLD     & 0.1660 & 0.0507    & \underline{0.0242} & 0.0254 & \textbf{0.0174} \\
                             & AKLD    & 0.2007 & 0.1504    & 0.1524 & \underline{0.1367} & \textbf{0.1283} \\ 
    \multirow{2}{*}{GP}      & KLD     & 0.1315 & 0.0781    & 0.0432 & \underline{0.0352} & \textbf{0.0143} \\ 
                             & AKLD    & 0.1968 & 0.1797    & 0.1273 & \underline{0.1180} & \textbf{0.1074} \\ 
    \multirow{2}{*}{IP}      & KLD     & 0.0581 & 0.5128    & 0.0410 & \underline{0.0339} & \textbf{0.0291} \\ 
                             & AKLD    & 0.2929 & 1.7490    & \underline{0.1145} & 0.1522 & \textbf{0.1128} \\ 
    \multirow{2}{*}{N6}      & KLD     & 0.3524 & 0.2026    & \underline{0.0206} & 0.0309 & \textbf{0.0167} \\ 
                             & AKLD    & 0.2919 & 0.2469    & 0.1304 & \underline{0.1108} & \textbf{0.1106} \\ 
    \multirow{2}{*}{S6}      & KLD     & 0.4517 & 0.3735    & 0.0302 & \underline{0.0272} & \textbf{0.0193} \\ 
                             & AKLD    & 0.4190 & 0.2641    & 0.1933 & \underline{0.1355} & \textbf{0.1223} \\ 
    \midrule[0.2pt]
    \multirow{2}{*}{Average} & KLD     & 0.2129 & 0.2435    & 0.0342 & \underline{0.0305} & \textbf{0.0194} \\
                             & AKLD    & 0.2802 & 0.5180    & 0.1436 & \underline{0.1306} & \textbf{0.1163} \\ 
    \bottomrule[0.5pt]
    \end{tabular}
    }
    \label{table:noise_gen_per_devices}
    \vspace{-2.5mm}
\end{table}

\begin{figure}[t]
\begin{center}
\centerline{\includegraphics[width=1.0\columnwidth]{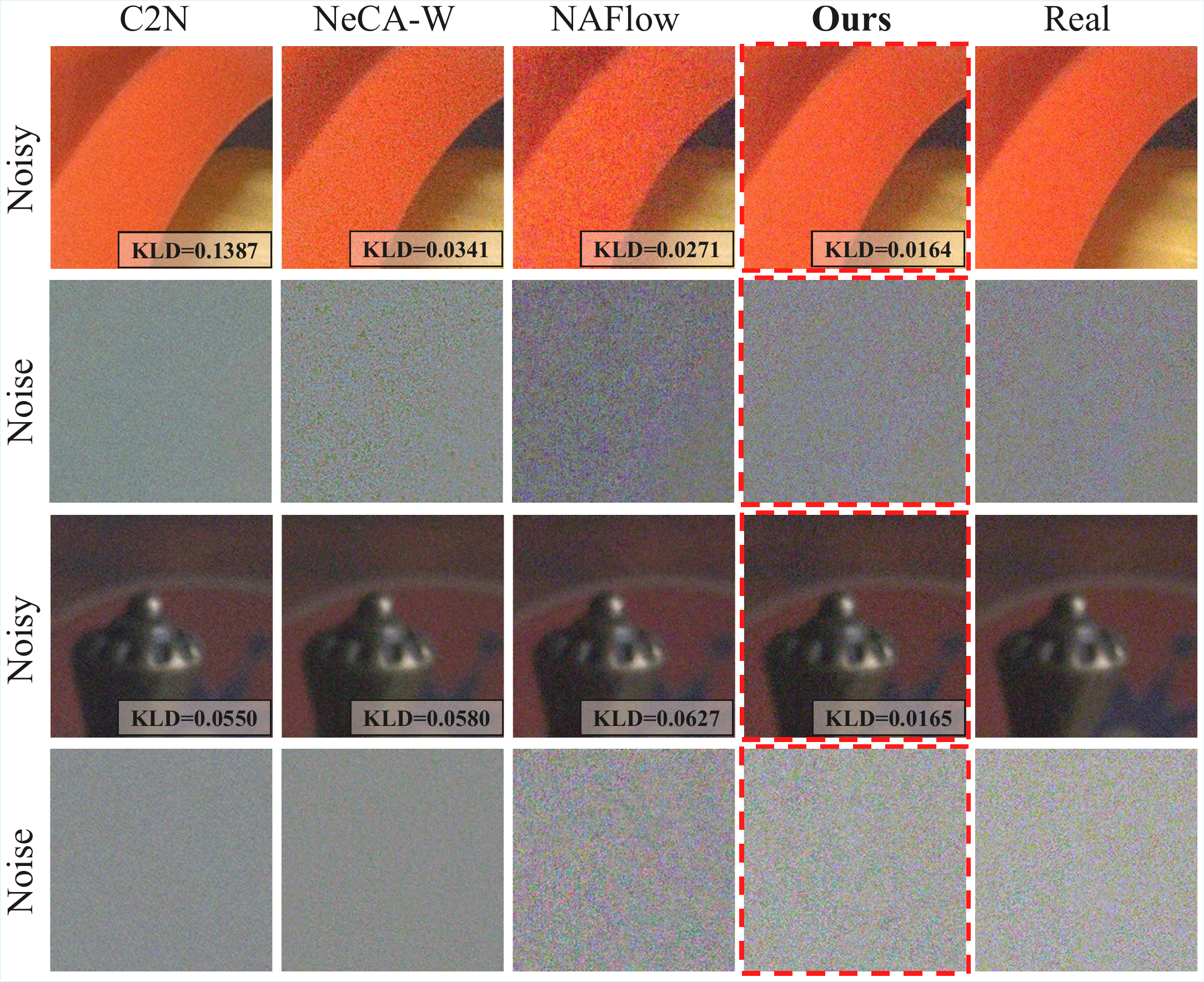}}
\caption{Visualization of synthetic noisy images on the SIDD validation set. From left to right: C2N, NeCA-W, NAFlow, Ours (\framework{}), and real noisy images.
}
\label{fig:noise_map}
\vspace*{-12.5mm}
\end{center}
\end{figure}

\noindent\textbf{Metrics.}
To examine the quality of the generated noise, we use two metrics: Kullback-Leibler Divergence (KLD) and Average Kullback-Leibler Divergence (AKLD)~\citep{danet}.
In addition, we employ PSNR and SSIM metrics to assess the performance of the downstream denoising network.

\noindent\subsection{Real-World sRGB Noise Generation and Removal} 
\label{sec:4.2} 
\noindent\textbf{Dataset.}
We employ the SIDD~\citep{SIDD} dataset, which includes 34 different camera configurations.
We adopt the SIDD Medium split, comprising 320 noisy-clean image pairs captured with five unique smartphone cameras: Google Pixel (GP), iPhone 7 (IP), Samsung Galaxy S6 Edge (S6), Motorola Nexus 6 (N6), and LG G4 (G4).
To evaluate the quality of the generated noise, we use the SIDD validation set. Additionally, SIDD+~\citep{siddplus}, PolyU~\citep{polyu}, and Nam~\citep{nam} 
are employed to further assess generality on both noise quality and denoising performance.

\noindent\textbf{Device-Specific Noise Quality Assessment.}
We train a single unified PNG on the full SIDD training set using all devices and evaluate the noise quality of each smartphone device type to assess device-specific noise generation performance in terms of KLD and AKLD.
As in \ttabref{table:noise_gen_per_devices}, our method is compared with four other models: C2N~\citep{c2n}, Flow-sRGB~\citep{Flow-sRGB}, NeCA-W~\citep{neca}, and NAFlow~\citep{naflow}.
Note that, to align with the experimental settings of previous works, 
we utilize some of the paired noisy-clean images from the SIDD validation set for noise generation, where the same metadata (\eg, smartphone manufacturer, ISO setting) is available for both the validation set and the training set.

Our \framework{} outperforms other methods across all device types, showing substantial improvements over NAFlow in both average KLD and AKLD scores, with improvements of 0.0111 and 0.0143, respectively.
We also present visual comparisons of the noisy images generated by each method in \ffigref{fig:noise_map}.
\framework{} generates more natural and realistic noise that closely resembles real-world noise distributions in terms of magnitude and correlation patterns, demonstrating its superior performance in generating realistic noise.

\begin{table}[t]
    \centering
    \caption{
    Denoising performance of DnCNN on SIDD-Benchmark in terms of PSNR$\uparrow$ and SSIM$\uparrow$.
    All methods are trained with synthetic noisy-clean pairs. Note that \textit{Real} indicates denoising results by training using real noisy-clean pairs. The best and second-best results are highlighted in \textbf{bold} and \underline{underline}.}
    \vspace{-2.5mm}
    \resizebox{1.0\columnwidth}{!}{
    \begin{tabular}{l|cccccc|c}
    \toprule[0.5pt]
     
        Metrics           & C2N   & NoiseFlow & Flow-sRGB & NeCA-W & NAFlow & \textbf{Ours} & \textit{Real} \\ 
        \midrule[0.2pt]
        PSNR$\uparrow$    & 33.76 & 33.81     & 34.74     & 36.82  & \underline{37.22}  & \textbf{37.55} & 37.63 \\
        SSIM$\uparrow$    & 0.901 & 0.894     & 0.912     & 0.932  & \underline{0.935}  & \textbf{0.937} & 0.936 \\ 
        \bottomrule[0.5pt]
    \end{tabular}
    }
    \label{table:sidd_denoising_performance}
    \vspace*{-2.5mm}
\end{table}

\begin{figure}[!t]
\begin{center}
\centerline{\includegraphics[width=1.0\columnwidth]{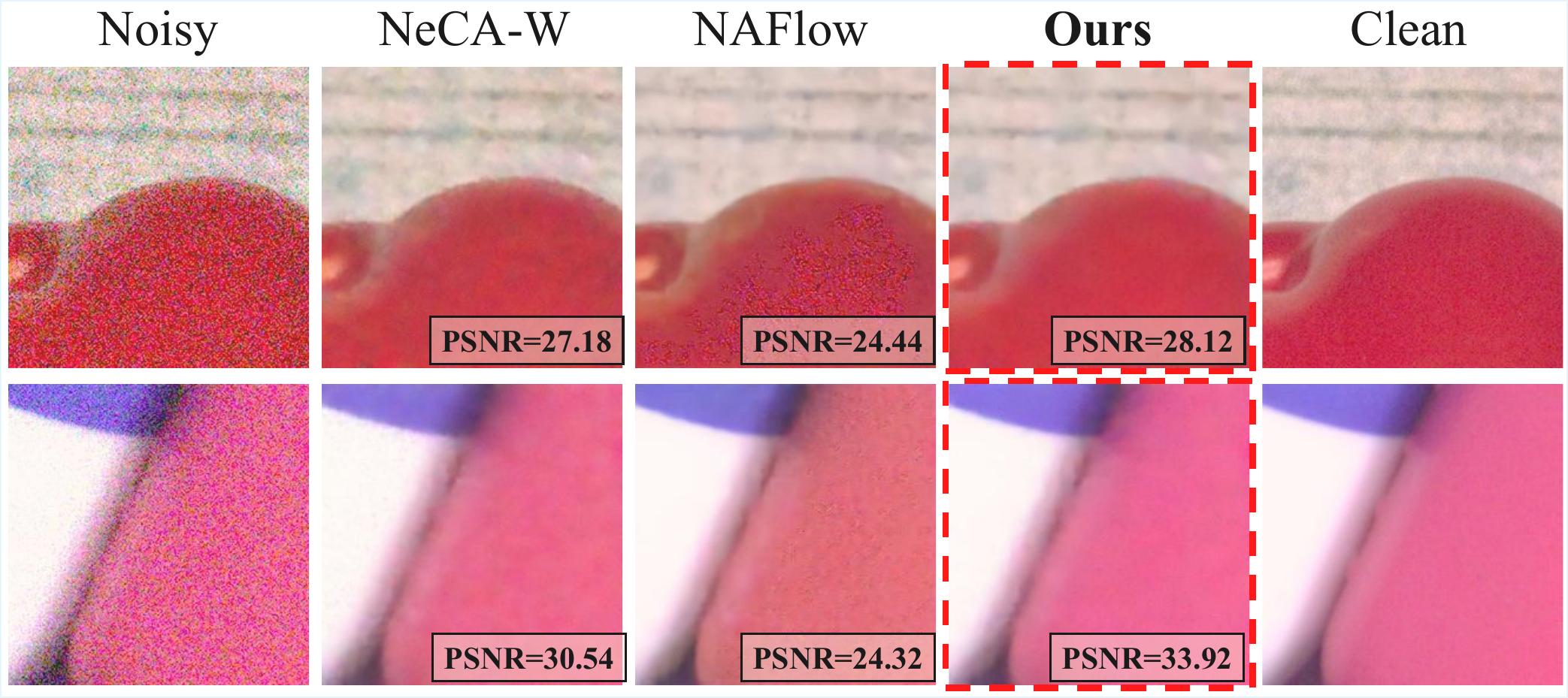}}
\caption{Visual comparison on denoising results with PSNR$\uparrow$ on SIDD validation set from DnCNN trained on each method. For more qualitative results, please refer to the supplementary material.}
\label{fig:denoising_results}
\vspace*{-12.5mm}
\end{center}
\end{figure}

\begin{table*}[t]
\centering
\caption{
Quantitative comparison of DnCNN denoising performance on the PolyU, Nam, SIDD validation, and SIDD+ benchmarks. 
The percentage (\%) denotes the mixing ratio between the two training subsets. The best and second-best results are highlighted in \textbf{bold} and \underline{underline}.
}
\vspace{-2.5mm}
\resizebox{0.9\textwidth}{!}{ 
\begin{tabular}{clcccccccc|cc}
\toprule[0.5pt] 
\multirow{2}{*}{Model} & \multirow{2}{*}{Methods}                    & \multicolumn{2}{c}{PolyU}                                        & \multicolumn{2}{c}{Nam}                                          & \multicolumn{2}{c}{SIDD Validation}                              & \multicolumn{2}{c}{SIDD+}                                                             & \multicolumn{2}{|c}{Average}                                      \\ \cmidrule(lr){3-12}  
                        &                                             & PSNR$\uparrow$                 & SSIM$\uparrow$                  & PSNR$\uparrow$                 & SSIM$\uparrow$                  & PSNR$\uparrow$                 & SSIM$\uparrow$                  & PSNR$\uparrow$                 & \multicolumn{1}{c|}{SSIM$\uparrow$}                  & PSNR$\uparrow$                 & SSIM$\uparrow$                  \\ \toprule[0.5pt]
\multirow{7}{*}{DnCNN}  & \textit{Real} (100\%)      & 36.34                          & 0.9204                          & 35.35                          & 0.8828                          & \underline{37.72} & 0.8905                          & 35.68                          & \multicolumn{1}{c|}{0.8860}                          & 36.27                          & 0.8949                          \\ \cmidrule(lr){2-12} 
                        & NeCA-W (100\%)                              & 36.96                          & 0.9544                          & 35.61                          & 0.9403                          & 36.26                          & 0.8812                          & 35.42                 & \multicolumn{1}{c|}{0.8900}                 & 36.06                          & 0.9165                          \\
                        & NAFlow (100\%)                              & 36.85                          & 0.9499                          & 37.45                          & 0.9524                          & 37.27                          & 0.8989                          & \textbf{36.67}                 & \multicolumn{1}{c|}{\textbf{0.9249}}                 & 37.06                          & 0.9315                          \\
                        & \textbf{Ours} (100\%)      & \underline{37.93} & \underline{0.9609} & \underline{38.08} & \textbf{0.9630}                 & 37.63                          & 0.8960                          & 36.23                          & \multicolumn{1}{c|}{0.9072}                          & \underline{37.47} & \underline{0.9318} \\ \cmidrule(lr){2-12} 
                        & NeCA-W-Mixed (50\%)                         & 37.92                          & 0.9601                          & 37.86                          & 0.9580                          & 37.51                          & 0.8952 & 35.79                          & \multicolumn{1}{c|}{0.8961}                          & 37.27                          & 0.9274                          \\
                        & NAFlow-Mixed (50\%)                         & 37.29                          & 0.9520                          & 37.47                          & 0.9538                          & 37.66                          & \underline{0.9043} & 36.27                          & \multicolumn{1}{c|}{0.9105}                          & 37.17                          & 0.9301                          \\
                        & \textbf{Ours-Mixed} (50\%) & \textbf{37.98}                 & \textbf{0.9610}                 & \textbf{38.09}                 & \underline{0.9617} & \textbf{37.96}                 & \textbf{0.9047}                 & \underline{36.57} & \multicolumn{1}{c|}{\underline{0.9137}} & \textbf{37.65}                 & \textbf{0.9353}                 \\ \toprule[0.5pt]
\end{tabular}
}
\label{table:denoising_qual_other_dataset_dncnn}
\vspace*{-5.0mm}
\end{table*}

\noindent\textbf{Denoising Results on SIDD.}
To assess the quality of noise generation methods, we train DnCNN denoising networks using generated noisy-clean paired SIDD training dataset and compare their denoising performance.
In \ttabref{table:sidd_denoising_performance}, we compare the DnCNN denoising performance of our~\framework{} with C2N, Flow-sRGB, NeCA-W, NAFlow, and \textit{Real}.
Specifically, \textit{Real} represents denoising results from real (not synthetic) noisy-clean paired data in the SIDD training set, serving as an oracle for synthetic noise generation.
Moreover, all noise generation methods are trained on the SIDD training dataset.  
\framework{} outperforms other methods and surpasses NAFlow by over 0.33 dB in PSNR and 0.002 in SSIM.
Notably, the denoising result produced by \framework{} closely matches that of \textit{Real}, achieving a minimal PSNR gap of only 0.08 dB and achieving a slightly higher SSIM by 0.001, indicating the high realism of the generated images.
For more details, the qualitative evaluation is depicted in \ffigref{fig:denoising_results}.

\noindent\textbf{Robustness on Various Real-World Datasets.}
To further assess denoising performance in real-world scenarios, we train DnCNN with our SIDD synthetic dataset and then evaluate it on four additional benchmarks: PolyU, Nam, SIDD validation, and SIDD+. 
These datasets cover a broad range of noise characteristics. \ttabref{table:denoising_qual_other_dataset_dncnn} compares our results with three baselines: NeCA-W and NAFlow, which are trained on synthetic noise, and \textit{Real}, which uses the ground-truth noisy-clean pairs from the SIDD training set.
We also report results from mixed training sets that combine equal amounts of real and synthetic data to further enhance performance and improve the practical applicability of synthetic data generation.

Denoising networks trained with our synthetic dataset achieve the best average performance in both the pure synthetic setting (100\%) and the mixed setting (50\%). In the pure setting, it matches or surpasses \textit{Real} on the smartphone-based SIDD and SIDD+ benchmarks and shows gains on the DSLR-based PolyU, and Nam datasets. 
These improvements are likely due to the larger scale and diversity of our generated noise, which helps prevent overfitting to a single domain.

In the mixed setting, performance further improves over the pure synthetic setting and exceeds \textit{Real} across all four benchmarks.
This demonstrates that our synthesized dataset enables the denoising network to effectively mitigate overfitting issues while also improving in-distribution performance. These results align with findings in other low-level vision dataset generation tasks, where synthesized datasets have been shown to alleviate overfitting problems due to their limited size of the training dataset~\cite{interflow, yang2023synthesizing, wang2021real, cai2019toward, degflow}.
Our findings suggest that our method is highly effective for real-world denoising tasks and provides a robust training dataset across diverse real-world imaging conditions.

\vspace{-2.5mm}
\subsection{Application: Metadata-Free Noise Generation}
To demonstrate the effectiveness of our approach, we perform an additional experiment in which metadata are absent or incompatible across datasets. As in the previous section, we compare both noise quality and denoising performance obtained with the synthesized data. 

\begin{table}[!t]
    \centering
    \caption{Quantitative results of synthetic noise on the PolyU, Nam, and MAI2021. All methods are trained with SIDD training set. The results are computed with KLD$\downarrow$ and AKLD$\downarrow$. The best results are shown in \textbf{bold}.}
    \resizebox{1.0\columnwidth}{!}{
        \begin{tabular}{lcccccc|cc}
        \toprule[0.5pt]
        \multirow{2}{*}{Methods} & \multicolumn{2}{c}{PolyU} & \multicolumn{2}{c}{Nam} & \multicolumn{2}{c}{MAI2021} & \multicolumn{2}{|c}{Average} \\ \cmidrule(lr){2-3} \cmidrule(lr){4-5} \cmidrule(lr){6-7} \cmidrule(lr){8-9} 
        \multicolumn{1}{r}{}                         & KLD$\downarrow$        & AKLD$\downarrow$      & KLD$\downarrow$        & AKLD$\downarrow$       & KLD$\downarrow$        & AKLD$\downarrow$       & KLD$\downarrow$        & AKLD$\downarrow$      \\ \midrule[0.2pt]
        Metadata-Dependent                                       & \xm             & \xm             & \xm             & \xm & \xm       & \xm             & \xm             & \xm       \\ \cmidrule(lr){1-9}
        NAFlow                                       & 0.2304          & 1.7801          & 0.2055          & 1.9320          & 0.7304          & 2.6465          & 0.3888          & 2.1195 \\
        \textbf{Ours}                                & \textbf{0.0822} & \textbf{0.5814} & \textbf{0.0966} & \textbf{0.4668} & \textbf{0.0747} & \textbf{0.3314} & \textbf{0.0845} & \textbf{0.4599} \\ \bottomrule[0.5pt]
        \end{tabular}
    }
    \label{table:noise_qual_other_dataset}
    \vspace{-0.5cm}
\end{table}

\noindent\textbf{Metadata-Free Noise Quality Assessment.}
In~\ttabref{table:noise_qual_other_dataset}, we validate the robustness of our method by evaluating the noise quality on external real-world datasets that were not used during the training phase.
We compare our~\framework{} with NAFlow trained on SIDD training dataset.
We evaluate using three external datasets: PolyU, Nam, MAI2021~\cite{mai2021} (no metadata or inconsistent metadata). These datasets were captured using various device types, camera sensors, and ISPs (\eg, smartphones, and DSLRs).
Note that metadata-dependent methods (\eg, NoiseFlow, Flow-sRGB,  NeCA-W) are unable to generate noisy images on external datasets where metadata is either unavailable or inconsistent with the metadata used during training.
To produce synthetic noisy images, real clean–noisy pairs of each external dataset are passed through \framework{} and NAFlow.
Notably, NAFlow and our method can synthesize noisy images using noisy-clean pairs from the external datasets without metadata. However, unlike our approach, NAFlow still requires metadata for training.
The overall results demonstrate that~\framework{} consistently outperforms NAFlow across all datasets, achieving substantial gains in average KLD and AKLD scores.
This emphasizes the robustness of \framework{} on external datasets across various camera settings, demonstrating its potential to expand dataset volume by generating noisy images from limited external datasets.

\begin{table}[!t]
    \centering
\caption{Quantitative denoising results on the external dataset.}
    \resizebox{1.0\columnwidth}{!}{
    \begin{tabular}{ccccc|cc}
    \toprule[0.5pt]
    \multirow{2}{*}{\begin{tabular}[c]{@{}c@{}}Test \\ Dataset\end{tabular}} & \multicolumn{2}{c}{\textit{Real} (100\%)} & \multicolumn{2}{c}{NAFlow-Mixed (50\%)} & \multicolumn{2}{|c}{\textbf{Ours-Mixed} (50\%)} \\ \cmidrule(lr){2-7} 
                             & PSNR$\uparrow$ & SSIM$\uparrow$ & PSNR$\uparrow$ & SSIM$\uparrow$ & PSNR$\uparrow$   & SSIM$\uparrow$           \\ \toprule[0.5pt]
    Nam                      & 38.87        & 0.9712       & 39.69        & 0.9715       & \textbf{40.40} & \textbf{0.9761}        \\ \bottomrule[0.5pt]
    \end{tabular}
    }
    \label{table:ood_noise_augmentation}
    \vspace{-3.5mm}
\end{table}

\noindent\textbf{Denoising Results on External Dataset.}
In \ttabref{table:ood_noise_augmentation}, we evaluate the generalization performance of our approach. 
Specifically, we use \framework{} trained on SIDD to augment an external dataset (NAM), then train a DnCNN network using this augmented data and evaluate the denoising performance.
For a fair comparison, we follow the same procedure using NAFlow trained on the same SIDD dataset.
The Nam dataset is divided into generation and evaluation subsets at a 4:1 ratio, and the real clean–noisy pairs in the generation subset are passed through \framework{}, to produce additional synthetic noisy images.
Moreover, \textit{Real} shows denoising results from training DnCNN solely on real datasets.
When combined with real data, our denoiser trained with 50\% real and 50\% synthetic data (Ours-Mixed) exceeds the \textit{Real} baseline by 1.53 dB in PSNR and 0.0049 in SSIM, outperforming NAFlow by 0.71 dB and 0.0046 on NAM evaluation subset. 
Overall, the results confirm that our unified framework can synthesize realistic noisy data from unseen domains, and incorporating these synthetic samples with real observations boosts denoising accuracy while reducing overfitting.

\vspace*{-2.5mm}
\subsection{Ablation Study}
In this section, we conduct ablation studies to evaluate our approach from two primary perspectives: (i) the impact of metadata classification, and (ii) the individual contributions of the GPB and LPB modules.
Due to space limitations, please refer to
~\ssecref{appendix:additional_ablation} 
for more comprehensive ablation studies.

\noindent\textbf{Metadata Classification}
\begin{table}[t]
    \centering
    \caption{Quantitative results of metadata classification on SIDD validation with different combinations of prompt features in terms of accuracy$\uparrow$. We test the validity of the prompt features in two aspects: camera sensor type and ISO level. The best and second-best results are highlighted in \textbf{bold} and \underline{underlined}.} 
    \resizebox{0.45\textwidth}{!}{
        \begin{tabular}{cccc}
        \toprule[0.5pt]
        \multirow{2}{*}{Classifier} & \multirow{2}{*}{\textit{Camera Sensor} (\%)} & \multicolumn{2}{c}{\textit{Camera Sensor + ISO Level}} \\ \cmidrule(lr){3-4} 
                               &                                & Top-1 (\%)            & Top-3 (\%) \\ \midrule[0.2pt]
        Baseline (No Prompt)               & 75.80                          & 66.27                 & 93.91 \\
        GPB             & \underline{82.37}                          & \underline{68.35}                 & \underline{94.71}  \\
        GPB + LPB            & \textbf{94.47}                 & \textbf{75.48}        & \textbf{98.64} \\ \bottomrule[0.5pt]
        \end{tabular}
    }
    \label{table:metadata_cls}
    \vspace*{-5.0mm}
\end{table}
We evaluate the extent to which the prompt features extracted from the Prompt Encoder $\mathcal{E}$ capture the characteristics of the input noise. 

First, we train ResNet-based classifiers~\cite{resnet} using either the input noise $\mathbf{n}_\mathrm{Real}$ or the prompt features as input, measuring their ability to accurately categorize camera sensor types (\ie, camera manufacturer).
We select five major camera sensors from the SIDD dataset, resulting in five distinct labels, denoted as \textit{Camera Sensor} in the second column of \ttabref{table:metadata_cls}.
The classification results show that using features of GPB as input improves camera sensor prediction accuracy compared to the baseline model, which uses real noise as input (without GPB and LPB).
Furthermore, combining GPB with LPB further enhances classification accuracy. 
These results demonstrate that the prompt features contain information about the camera sensor.

Next, we create a total of 16 labels by simultaneously considering these five sensors and additional ISO levels, and conduct classification experiments.
We compare their classification performances on the SIDD validation set, measuring Top-1 and Top-3 accuracy (last two columns in \ttabref{table:metadata_cls}).
The results show that using both GPB and LPB achieves the best classification performance for identifying the camera sensor and ISO configuration, highlighting the effectiveness of the proposed prompt blocks in capturing input noise-dependent characteristics.

\begin{table}[t]
    \centering
    \caption{Effect of GPB and LPB on SIDD validation noise generation. The best results are shown in \textbf{bold}.} 
    \resizebox{0.5\columnwidth}{!}{
        \begin{tabular}{cccc}
            \toprule[0.5pt]
            \multicolumn{2}{c}{Combination} & \multicolumn{2}{c}{Generation} \\ \cmidrule(lr){1-2} \cmidrule(lr){3-4}
            GPB & LPB & KLD$\downarrow$ & AKLD$\downarrow$ \\ \midrule[0.2pt]
            \xm   & \xm   & 0.6182           & 0.4387 \\
            \cm   & \xm   & 0.0287          & 0.1112 \\
            \cm   & \cm   & \textbf{0.0261} & \textbf{0.1108} \\ \bottomrule[0.5pt]
        \end{tabular}
    }
    \label{table:ablation_prompt_block_on_noise_generation}
    \vspace*{-4.0mm}
\end{table}

\noindent\textbf{Effect of GPB and LPB on Noise Generation.}
To evaluate the impact of the prompt features extracted by the GPB and LPB, we conducted noise generation experiments, as shown in~\ttabref{table:ablation_prompt_block_on_noise_generation}.
The P-DiT model is differently trained with various variants of pretrained PAE to investigate the effect of each prompt feature.
Combining diverse information from the prompt blocks such as ISO settings and noise correlation enhances the quality of synthesized noise by providing vital input-specific distribution details.

\vspace{-1mm}
\section{Conclusion}
\vspace{-1mm}
In this work, we propose a novel~\framework{} framework to remove the reliance on metadata when generating real-world noise.
Our \framework{} comprises two key components: Prompt Autoencoder (PAE) and Prompt DiT (P-DiT).
The PAE encodes input noise, generating a compact latent code and extracting meaningful input-dependent prompt features at various scales to replace input noise-specific metadata.
The P-DiT then synthesizes latent codes using the prompt features and clean images as conditions. 
Finally, the generated latent codes are fed into the PAE Decoder alongside clean images to produce realistic signal-dependent noisy images.
A key advantage of our method is that it does not require explicit metadata during either the training or testing phases, unlike conventional methods, highlighting its strength in real-world applications.
Moreover, experimental results on real-world benchmarks demonstrate that our framework excels in both real-world noise generation and downstream denoising tasks.

\clearpage
\section*{Acknowledgments}
This work was supported by 
Institute of Information \& communications Technology Planning \& Evaluation (IITP) 
grant funded by the Korea government (MSIT) (No. RS2023-00220628, Artificial intelligence for prediction of structure-based protein interaction reflecting physicochemical principles), IITP grant funded by the Korea government (MSIT) (No.RS-2020-II201373, Artificial Intelligence Graduate School Program [Hanyang University]) and the research fund of Hanyang University (HY-2026).

{
    \small
    \bibliographystyle{ieeenat_fullname}
    \bibliography{main}
}

\clearpage
\setcounter{page}{1}
\maketitlesupplementary

\setcounter{table}{0}
\renewcommand{\thetable}{S\arabic{table}}
\setcounter{figure}{0}
\renewcommand{\thefigure}{S\arabic{figure}}
\setcounter{equation}{0}
\renewcommand{\theequation}{S\arabic{equation}}

\setcounter{section}{18}
\renewcommand{\thesection}{\Alph{section}}

\tableofcontents

\section{Supplementary Material}
\label{sec:Appendix}

\subsection{Prompt DiT (P-DiT)}
\label{sec_pdit}

\begin{figure*}[t]
\begin{center}
\centerline{\includegraphics[width=0.9\textwidth]{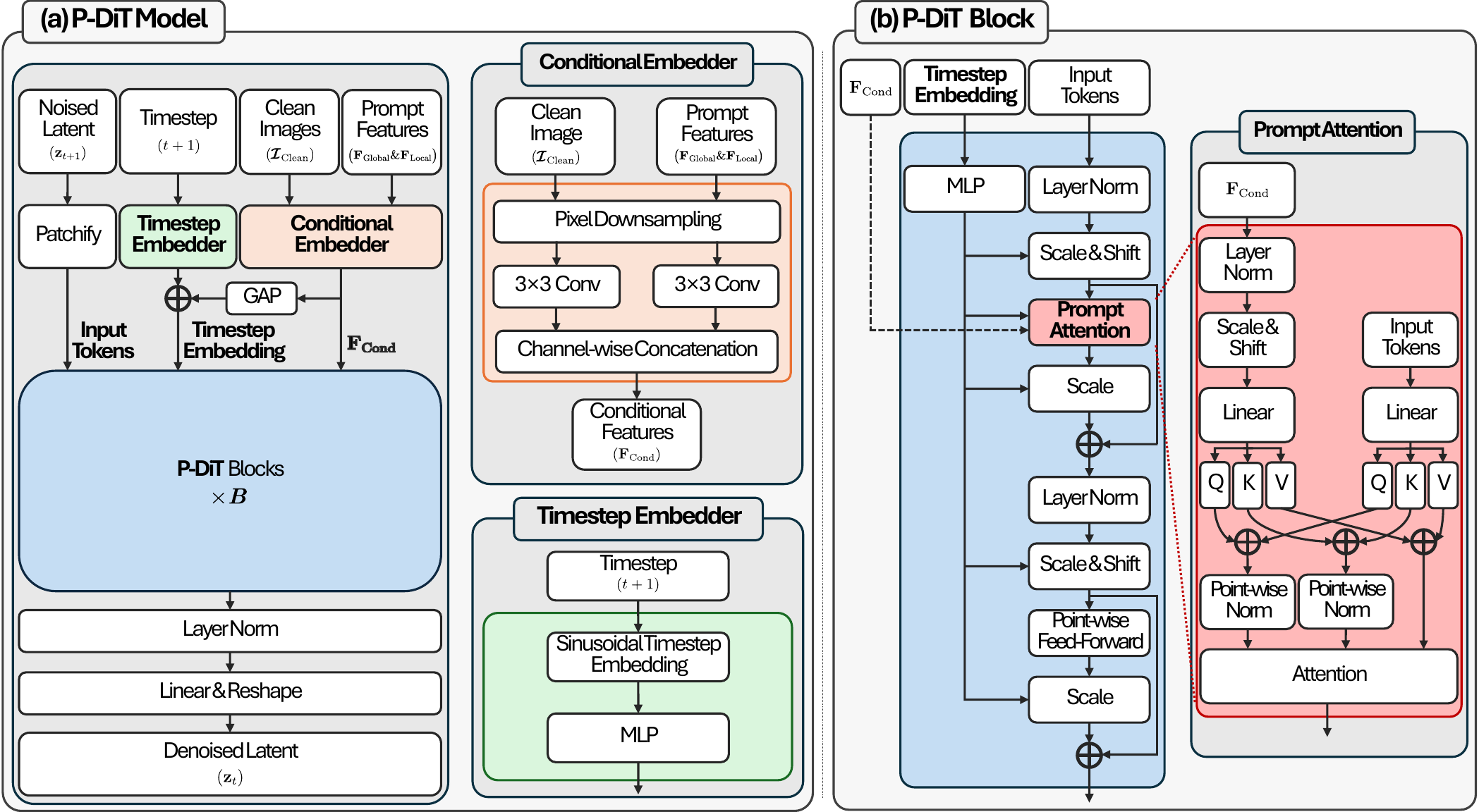}}
\caption{Overview of Prompt DiT (P-DiT). (a) Overall P-DiT structure. (b) P-DiT block.}
\label{fig:arch_pdit}
\vspace{-10mm}
\end{center}
\end{figure*}

In~\ffigref{fig:overall_flow} and~\ffigref{fig:arch_pdit}, we introduce P-DiT, which fully leverages the prompt features extracted from the Prompt Encoder $\mathcal{E}$ to synthesize latent codes that align with the embedded information of the input noise characteristics.
Our P-DiT is based on DiT~\citep{dit}, a transformer-based CM architecture specifically designed for training diffusion models.

In~\ffigref{fig:arch_pdit} (a), the P-DiT consists of a series of $B$ P-DiT blocks through which the noise-added input latent $\mathbf{z}_{t+1}$ is processed. 
Moreover, for conditioning features, we use the timestep $t+1$, clean images \text{\footnotesize $\pmb{\mathcal{I}}_\mathrm{Clean}$}, and prompt features \text{\footnotesize $\mathrm{\mathbf{F}}_\mathrm{Local}$} and \text{\footnotesize $\mathbf{F}_\mathrm{Global}$}. The timestep enables the model to adjust its denoising predictions based on the noise levels of the diffusion process at each timestep. 
Precisely, the timestep is embedded by a timestep embedder, composed of sinusoidal embeddings and an MLP block~\citep{attention}.
We also use the clean image to help the model learn the signal-dependent properties of real-world noise and prompt features to learn the camera-dependent properties. This enables the generation of latent codes that capture input noise characteristics without relying on metadata, with these features embedded by a conditional embedder.
Specifically, we first downsample these inputs to match the spatial size of the latent codes and then extract shallow features separately using $3{\times}3$ convolutional layers, which are concatenated along the channel axis as follows:
\begin{equation}
\begin{gathered}
\mathrm{\mathbf{F}}_\mathrm{Cond}=\Big\llbracket\mathrm{Conv}_{3\times3}\big(\mathrm{PD}(\pmb{\mathcal{I}}_\mathrm{Clean})\big), \mathrm{Conv}_{3\times3}\big(\mathrm{PD}(\mathrm{\mathbf{F}}_\mathrm{Local})\big), \\ \mathrm{Conv}_{3\times3}\big(\mathrm{PD}(\mathbf{F}_\mathrm{Global})\big)\Big\rrbracket,
\end{gathered}
\end{equation}
where $\mathrm{PD}(\cdot)$ indicates a pixel downsampling operator. Note that we use \text{\footnotesize $\mathbf{F}_\mathrm{Global}$} for all scale levels, and each feature is processed separately.
The concatenated feature \text{\footnotesize $\mathrm{\mathbf{F}}_\mathrm{Cond}$} is then processed with global average pooling and added to the time embeddings.

In \ffigref{fig:arch_pdit} (b), we present the P-DiT block. 
We employ adaptive layer normalization (AdaLN) as the conditioning mechanism to modulate the statistics of input features.
AdaLN consists of two components: layer normalization~\citep{layernorm} and adaptive modulation. 
AdaLN first normalizes the input features to have a mean of zero and a standard deviation of one.
Then, the normalized features are modulated using scale and shift parameters derived from the conditioning input.

Moreover, to fully leverage the available information from the conditions and capture locally varying spatial correlations characteristics in prompt features, we further enhance these features through a prompt attention mechanism, referred to as \textbf{Prompt Attention}.

\noindent\textbf{Prompt Attention.}
As illustrated in \ffigref{fig:arch_pdit} (b), we use the conditions to generate the key (K), query (Q), and value (V) within the attention layer by projections, similar to MMDiT~\citep{mmdit},
enabling the model to effectively capture the spatial information of the prompt features.
Specifically, we first modulate the combined conditional features \text{\footnotesize $\mathbf{F}_{\text{Cond}}$} through AdaLN to extract time-dependent information. 
Then, we generate the key , query, and value features using a single linear layer and combine them with the corresponding features from the input tokens through element-wise addition. 
Using these conditioned features, we apply cosine attention~\citep{cosinenorm, edm2} with point-wise normalization~\citep{pggan} to stabilize the training process~\citep{edm2}.

\subsection{Training details of P-DiT}
\label{appendix:training_details}
\subsubsection{CM Parameterization}
The CM-based model $f_{\theta}$ aims to approximate the consistency function $f(\cdot, \cdot)$, which satisfies $f(\mathbf{x}_{t}, \sigma_t) = \mathbf{x}_0$. Therefore, it must adhere to the boundary condition $f(\mathbf{x}_0, \sigma_0) = \mathbf{x}_0$. To ensure this, we follow the parameterization used in EDM~\citep{edm} and CM~\citep{cm}, defining the model as follows:
\begin{equation}
    f_{\theta}(\mathbf{x}_t, \sigma_t) = c_{\text{skip}}(\sigma_t)\mathbf{\mathbf{x}_t} + c_{\text{out}}(\sigma_t)F_{\theta}(c_{\text{in}}(\sigma_t)\mathbf{x}_t, \sigma_t),
\label{eq:edm_formulation}
\end{equation}
where $F_{\theta}$ is a free-form neural network, such as P-DiT, and $c_{\text{in}}$, $c_{\text{out}}$, and $c_{\text{skip}}$ control the scaling of input, output magnitudes, and the skip connection, respectively. These can be expressed as:
\begin{equation}
\begin{gathered}
c_{\text{in}}(\sigma_t) = \frac{1}{\sqrt{\sigma_{\text{data}}^2 + \sigma_t^2}}, \quad
c_{\text{skip}}(\sigma_t) = \frac{\sigma_{\text{data}}^2}{(\sigma_t - \sigma_0)^2 + \sigma_{\text{data}}^2}, \\
c_{\text{out}}(\sigma_t) = \frac{\sigma_{\text{data}}(\sigma_t - \sigma_0)}{\sqrt{\sigma_{\text{data}}^2 + \sigma_t^2}}.
\end{gathered}
\end{equation}
These formulations meet the boundary conditions where $c_{\text{skip}}(\sigma_0) = 1$ and $c_{\text{out}}(\sigma_0) = 0$.

\subsubsection{CM Hyperparameter} 
We detail the hyperparameters for training P-DiT, adopting most from iCT~\citep{ict} for CM.

\noindent\textbf{Discretization Curriculum.}
The discretization curriculum is designed to enhance CM training by systematically increasing the number of discretization timesteps $T$ in~\eeqref{eq:1}, 
improving the quality of generated samples. The discretization curriculum $\mathcal{C}(k)$ is defined as follows:
\begin{equation}
\mathcal{C}(k) = \min(s_0 2^{\frac{k}{K'}}, s_1) + 1, \quad \text{where} \ K' = \left\lfloor \frac{K}{\log_2\frac{s_1}{s_0}+ 1} \right\rfloor.
\end{equation}
In this context, $k$ ranges over $\{0, 1, \dots, K\}$, where $K$ denotes the total number of training iterations. The parameters $s_0$ and $s_1$ refer to the minimum and maximum discretization steps, respectively.
While iCT~\citep{ict} uses $s_0 = 10$ and $s_1 = 1280$, we empirically found that setting maximum number of discretization steps $s_1$ to 160 is sufficient to produce competitive performance.

\noindent\textbf{Noise Schedule.}
The noise schedule plays a crucial role in determining the sampling of noise levels during the CM training, significantly influencing the quality of the generated samples.
To define the noise schedule, we first discretize the noise level as follows: $\sigma_{\text{min}} = \sigma_{0} < \sigma_{1} < \dots < \sigma_{T} = \sigma_{\text{max}}$ where $\sigma_{\text{min}} = 0.002, \ \sigma_{\text{max}} = 80$.
As in~\citep{edm, cm, ict}, we set $\sigma_t$ as: 
\begin{equation}
\sigma_t = \left(\sigma_{\text{min}}^{1/\tau} + \frac{t-1}{\mathcal{C}(k)-1}\left(\sigma_{\text{max}}^{1/\tau} - \sigma_{\text{min}}^{1/\tau}\right)\right)^{\tau},
\end{equation}
where $\ t \in \{1, 2, \dots, \mathcal{C}(k)\}$. We use $\tau = 7$ and $\tau$ controls for the step length between noise levels $\sigma_t$ and $\sigma_{t+1}$. As $\tau$ increases, the step length at lower noise levels decreases, allowing the model to better capture high-frequency details.

Additionally, we utilize a lognormal distribution for noise level sampling, which reduces the emphasis on higher noise levels and mitigates the accumulation of errors in CT loss at lower noise levels. The noise sampling schedule is defined as:
\begin{equation}
\begin{gathered}
\sigma_t, \ \text{where} \ t \sim p(t), \ \text{and} \\ \ p(t) \propto \text{erf}\left(\frac{\log(\sigma_{t+1}) - P_{\text{mean}}}{\sqrt{2} P_{\text{std}}}\right) - \text{erf}\left(\frac{\log(\sigma_t) - P_{\text{mean}}}{\sqrt{2} P_{\text{std}}}\right),
\end{gathered}
\end{equation}
where erf indicates error function, and $P_{\text{mean}}, P_{\text{std}}$ determine the shape of log distribution. We choose $P_{\text{mean}} = -1.1, \ P_{\text{std}} = 2.0$, following iCT~\citep{ict}.

\noindent\textbf{Loss Function.}
In~\eeqref{eq:ct_loss}, we use the pseudo-Huber loss~\citep{ict} as the distance function $d(\cdot)$. The pseudo-Huber loss transitions between $\mathcal{L}_{1}$ and $\mathcal{L}_{2}$ metrics and is more robust to outliers than the $\mathcal{L}_{2}$ metric. The pseudo-Huber loss is defined as:
\begin{equation}
d(\mathbf{x}, \mathbf{y}) = \sqrt{\|\mathbf{x} - \mathbf{y}\|_2^2 + c^2} - c,
\end{equation}
where $c = 0.00054 \sqrt{m}$, and $m$ indicates data dimensionality.

\noindent\textbf{Loss Weighting.}
The weighting function $\lambda(\cdot)$ in~\eeqref{eq:ct_loss} modulates the significance of CT losses across varying noise levels during training. Following~\citep{ict}, we set the weighting function $\lambda(\cdot)$ as follows:
\begin{equation}
\lambda(\sigma_t) = \frac{1}{\sigma_{t+1} - \sigma_t}.
\end{equation}
By assigning lower weights to higher noise levels, the weighting function ensures that the model focuses on learning from lower noise levels, where the data is more distinct and informative. This strategy improves sample quality by reducing the influence of errors associated with higher noise levels, thereby enhancing the overall performance of consistency models.

\subsubsection{Latent Code Normalization}
The formulation of EDM~\cite{edm} assumes that the mean and standard deviation of the training data are zero and $\sigma_{\text{data}}$, respectively, as stated in~\eeqref{eq:edm_formulation}. Following the approach in~\citep{edm2}, we also normalize the encoded latent codes using precomputed statistics from the training data. Specifically, we first calculate the channel-wise mean and standard deviation of the latent codes from the training dataset. Then, we subtract the input latent code by the precomputed mean to achieve a mean of zero, and divide it by the precomputed standard deviation, followed by multiplying by $\sigma_{\text{data}}$, to set the standard deviation to $\sigma_{\text{data}}$. When the latent codes are generated, we reverse this procedure before transforming them back to the image space via the Decoder $\mathcal{D}$.

\subsubsection{P-DiT Hyperparamters}
Following the approach in~\citep{cm, ict}, we update P-DiT parameters using an exponential moving average with a decay rate of 0.9999 to stabilize the training process. 
P-DiT's model hyperparameters are based on DiT-S~\citep{dit}, except for the number of blocks (\(B = 8\)), to improve efficiency. The input noised latent is tokenized with a patch size of 1, allowing for finer noise information to be embedded in the latent code. 
To mitigate overfitting, we apply a dropout rate of 0.1 to the pointwise feed-forward layer and add minor noise and apply downsampling operation with a factor of 2 to the conditional clean images~\citep{cm}.

\begin{table}[t]
\centering
\caption{Inference speed comparison between NeCA-W, NAFlow, and \framework{}.}
\vspace{0.1cm}
\resizebox{0.35\textwidth}{!}{
\begin{tabular}{ccc|c}
        \toprule[0.5pt]
        \multirow{2}{*}{Resolution} & \multicolumn{3}{c}{Inference Time (image / second)} \\ \cmidrule(lr){2-4} 
                             & NeCA-W & NAFlow & \textbf{Ours} \\ \midrule[0.2pt]
        $256{\times}256$     & 150    & 13     & \textbf{57} \\
        $512{\times}512$     & 38     & 8      & \textbf{21} \\
        $1024{\times}1024$   & 9      & 4      & \textbf{5} \\ \bottomrule[0.5pt]
        \end{tabular}
    }
    \label{table:computational_complexity}
\vspace*{-3mm}
\end{table}

\subsection{Model Size and Inference Speed of \framework{}}
\label{appendix:model_capacity}
Our \framework{} consists of two sub-models, PAE with 14.9M parameters and P-DiT with 29.1M, yielding roughly 44M parameters in total. This total is comparable to NeCA-W \citep{neca}, whose camera-specific model for each of the five manufacturers in the SIDD dataset contains 40.5M parameters.
To train both PAE and P-DiT, we utilize four NVIDIA A6000 GPUs.
Following the approach in~\citep{cm, ict, edm, edm2}, we train P-DiT using mixed-precision, which reduces both training time and memory consumption.

To evaluate the computational complexity of our two-stage framework, we compare the inference time at different image resolutions with two other SOTA models, NeCA-W~\cite{neca} and NAFlow~\cite{naflow}. Specifically, we measure the number of images that each model can process per second on a single A6000 GPU.
As shown in \ttabref{table:computational_complexity}, with a resolution of $256{\times}256$, \framework{} can generate 57 images per second, which is about 4.38 times faster than NAFlow. At a $512{\times}512$ resolution, \framework{} generates 21 images per second, 2.625 times faster than NAFlow. At the highest resolution of $1024{\times}1024$, \framework{} still performs 1.25 times faster than NAFlow, producing 5 images per second, demonstrating its practicality at high image resolutions.
Even though NeCA-W shows the highest speed at all resolutions, the inference time difference ratio between NeCA-W and \framework{} decreases as the resolution increases, from 2.63 to 1.8. Hence, our method demonstrates strong performance while maintaining reasonable computational complexity at high resolutions compared to recent SOTA models.

\begin{table}[!t]
\centering
        \caption{Effect of conditioning features on different components in P-DiT. The best results are shown in \textbf{bold}.}
        \vspace{0.1cm}
        \resizebox{0.7\columnwidth}{!}{
        \begin{tabular}{cccc}
        \toprule[0.5pt]
        \multicolumn{2}{c}{Conditioning} & \multicolumn{2}{c}{Generation}    \\ \cmidrule(lr){1-2} \cmidrule(lr){3-4}
        Timestep           & Attention          & KLD↓            & AKLD↓           \\ \midrule[0.2pt]
        \xm                & \xm                & 0.5661          & 0.4132          \\
        \cm                & \xm                & 0.0287          & 0.1291          \\
        \cm                & \cm                & \textbf{0.0261} & \textbf{0.1108} \\ \bottomrule[0.5pt]
        \end{tabular}}
        \label{tab:ablation_cond_feat}
\end{table}

\begin{table}[!t]
\centering
\caption{Effect of GPB and LPB on SIDD validation noisy image reconstruction. The best results are shown in \textbf{bold}.}
\vspace{0.1cm}
\resizebox{0.60\columnwidth}{!}{
\begin{tabular}{cccc}
    \toprule[0.5pt]
    \multicolumn{2}{c}{Combination} & \multicolumn{2}{c}{Reconstruction} \\ \cmidrule(lr){1-2} \cmidrule(lr){3-4}
    GPB & LPB & PSNR$\uparrow$ & SSIM$\uparrow$ \\ \midrule[0.2pt]
    \xm   & \xm   & 37.58 & 0.9800 \\
    \cm   & \xm   & 46.30 & 0.9982 \\
    \cm   & \cm   & \textbf{46.54} & \textbf{0.9983} \\ \bottomrule[0.5pt]
    \end{tabular}
    }
    \label{table:ablation_prompt_block_on_noisy_image_reconstruction}
\end{table}

\subsection{Additional Ablation Studies}
\label{appendix:additional_ablation}
\subsubsection{Effect of Conditioning Features in P-DiT}
P-DiT generates latent codes that embed noise information through conditional features. Specifically, in \ffigref{fig:arch_pdit} (a), we first add the conditional feature $\mathbf{F}_{\text{Cond}}$ to the timestep embedding, and in \ffigref{fig:arch_pdit} (b), we condition the conditional feature through prompt attention. 

In \ttabref{tab:ablation_cond_feat}, we investigate the effect of conditional features on two different components: the timestep embedding and attention.
P-DiT without any conditional features fails to capture the target noise information, resulting in inferior performance in terms of KLD and AKLD. The model that conditions the conditional features on both components demonstrates superior performance compared to the model that only conditions them on the timestep embedding. This highlights that utilizing spatial information through the proposed prompt attention allows the model to fully exploit the conditional features.

\subsubsection{Effect of GPB and LPB on Noisy Image Reconstruction}
In~\ttabref{table:ablation_prompt_block_on_noisy_image_reconstruction}, we also assess the influence of both prompt blocks on noisy image reconstruction.
Incorporating GPB and LPB enhances the fidelity of the reconstructed images, as the prompt features encompass meaningful information about noise characteristics at different scales (see~\ffigref{fig:arch_pe}). 
Therefore, it is essential to utilize both prompt features to enhance the quality of the reconstructed images.
In particular, we pass the latent directly from the Prompt Encoder $\mathcal{E}$ to the Decoder $\mathcal{D}$, without using P-DiT for this experiment.

\subsubsection{Effect of Number of P-DiT Blocks}
As shown in~\ttabref{table:cmp_num_blocks}, we measure the KLD scores across different numbers of blocks $B$ in the P-DiT to evaluate their impact on noise generation, with $B$ representing the hierarchical levels used in our P-DiT.
The results indicate that increasing the number of blocks improves the quality of noise generation.
Even with half the number of blocks, P-DiT yet achieves a moderate KLD score, with only a slight difference from the final version ($B$=8), suggesting an opportunity to optimize the trade-off between model complexity and performance.

\begin{table}[!t]
\centering
\caption{KLD score depending on different number of blocks $B$.} 
\resizebox{0.35\columnwidth}{!}{
\begin{tabular}{cc} 
    \toprule[0.5pt]
    \# Blocks & KLD$\downarrow$  \\ 
    \midrule[0.2pt]
    $B=4$    & 0.0350                   \\
    $B=6$    & 0.0345                   \\
    $B=8$    & \textbf{0.0261}                   \\
    \bottomrule[0.5pt]
\end{tabular}
}
\label{table:cmp_num_blocks}
\end{table}

\begin{figure*}[ht]
\begin{center}
\centerline{\includegraphics[width=0.9\textwidth]{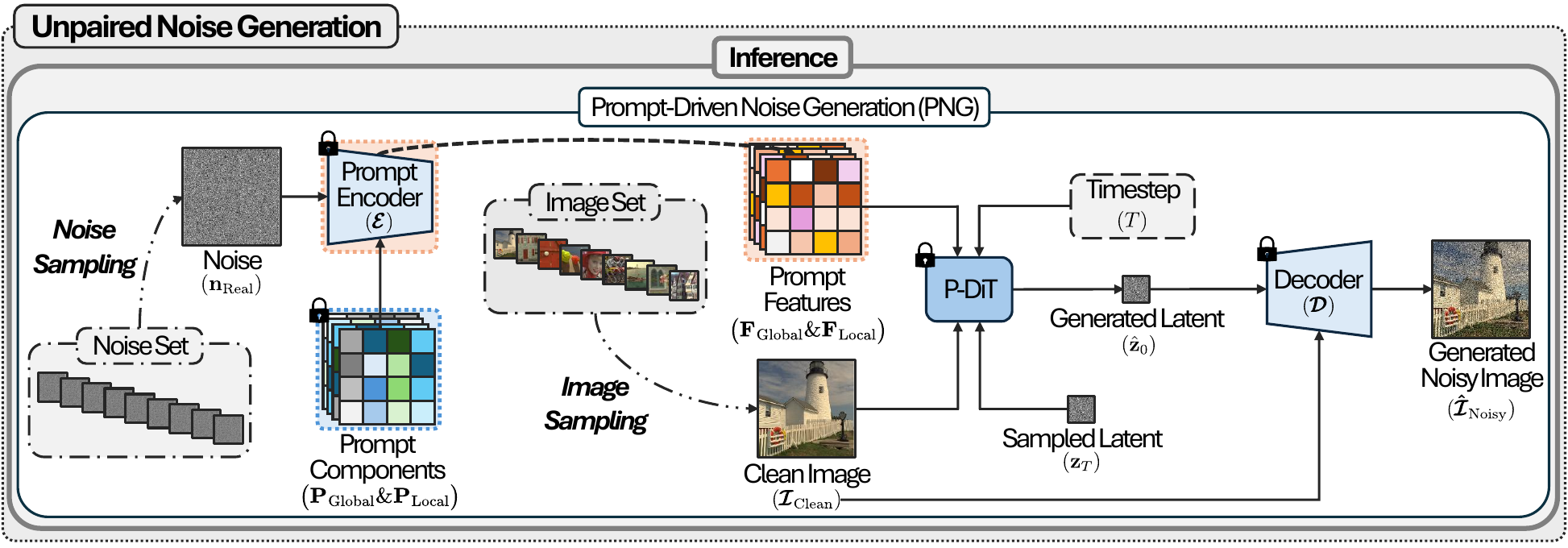}}
\caption{Overview of unpaired noise generation process.}
\label{fig:appx_unpaired_noise_generation}
\vspace*{-7.5mm}
\end{center}
\end{figure*}

\begin{table*}[ht]
\centering
\caption{Quantitative results on denoising generalization performance depending on the size of the synthesized dataset using \framework{}.
The multiplication sign ($\times$) indicates the scaling applied to the original number of patches. The best and second-best results are denoted as \textbf{bold} and \underline{underline}, respectively.}
\vspace{0.2cm}
\resizebox{0.80\textwidth}{!}{
\begin{tabular}{ccccccccc|cc}
\toprule[0.5pt]
\multirow{2}{*}{\# Samples} & \multicolumn{2}{c}{PolyU} & \multicolumn{2}{c}{Nam}  & \multicolumn{2}{c}{SIDD Validation} & \multicolumn{2}{c}{SIDD+} & \multicolumn{2}{|c}{Average} \\ \cmidrule(lr){2-3} \cmidrule(lr){4-5} \cmidrule(lr){6-7} \cmidrule(lr){8-9} \cmidrule(lr){10-11}
                         & PSNR$\uparrow$          & SSIM$\uparrow$           & PSNR$\uparrow$          & SSIM$\uparrow$           & PSNR$\uparrow$          & SSIM$\uparrow$           & PSNR$\uparrow$         & SSIM$\uparrow$           & PSNR$\uparrow$          & SSIM$\uparrow$  \\ 
                         \midrule[0.2pt]
$\times$1                   & 37.40             & \textbf{0.9569}    & 37.29             & 0.9578             & 36.55             & \underline{0.8887} & 35.72             & 0.8997             & 36.74 & 0.9258 \\
$\times$2                   & \underline{37.61} & \underline{0.9546} & \underline{37.92} & \textbf{0.9580}    & \underline{36.95} & 0.8838             & \underline{35.87} & \textbf{0.9078}    & \underline{37.09} & \underline{0.9260} \\
$\times$4                   & \textbf{37.72}    & 0.9542             & \textbf{37.94}    & \underline{0.9578} & \textbf{37.27}    & \textbf{0.8955}    & \textbf{36.00}    & \underline{0.9022} & \textbf{37.23} & \textbf{0.9274} \\ 
\bottomrule[0.5pt]
\end{tabular}
}
\label{table:dataset_expansion}
\vspace{-10px}
\end{table*}

\subsection{Further Analysis of the Generalization Performance}
\label{appendix:denoising_generalization_performance}
As shown in~\ttabref{table:denoising_qual_other_dataset_dncnn}, denoising networks trained with our synthesized dataset exhibit strong robustness to unseen noise. We further conduct experiments to assess the generalization capability of our approach on external datasets. The results demonstrate that expanding the training dataset through synthetic image generation helps mitigate overfitting to the training distribution and leads to improved performance on external dataset.

For the experiments, we select 15,000 non-overlapping patches from the SIDD training set to generate synthetic noisy images to train the DnCNN network. This setup ensures that each noisy sample is unique, providing a controlled environment to systematically analyze the impact of the size of the dataset on generalization.

In~\ttabref{table:dataset_expansion}, we organize the training data into three distinct groups: $\times$1, $\times$2, and $\times$4, where each factor indicates the number of synthetic noisy samples generated per clean image using \framework{}. Importantly, performance on the in-distribution dataset (SIDD+) remains consistent across all groups. However, the denoising network’s robustness on external datasets improves as the number of training samples increases, with the $\times$4 group achieving the highest average performance. These findings suggest that increasing noise diversity through synthetic data generation effectively reduces overfitting and highlights the efficacy of the \framework{} approach.

\begin{table*}[t]
\centering
\caption{Quantitative results for the noise generation performance on SIDD validation set under unpaired settings. The results are reported using KLD$\downarrow$ and AKLD$\downarrow$, with the best and second-best results highlighted in \textbf{bold} and \underline{underline}, respectively. \textsuperscript{\textdagger} indicates that the model is evaluated under an unpaired setting.}
\vspace{0.1cm}
\resizebox{0.9\textwidth}{!}{
\begin{tabular}{lcccccccccc|cc}
\toprule[0.5pt]
\multicolumn{1}{c}{\multirow{2}{*}{Methods}} & \multicolumn{2}{c}{G4} & \multicolumn{2}{c}{GP} & \multicolumn{2}{c}{IP} & \multicolumn{2}{c}{N6} & \multicolumn{2}{c}{S6} & \multicolumn{2}{|c}{Average} \\ \cmidrule(lr){2-3} \cmidrule(lr){4-5} \cmidrule(lr){6-7} \cmidrule(lr){8-9} \cmidrule(lr){10-11} \cmidrule(lr){12-13} 
\multicolumn{1}{c}{}                         & KLD\ensuremath{\downarrow}        & AKLD\ensuremath{\downarrow}      & KLD\ensuremath{\downarrow}        & AKLD\ensuremath{\downarrow}      & KLD\ensuremath{\downarrow}        & AKLD\ensuremath{\downarrow}      & KLD\ensuremath{\downarrow}        & AKLD\ensuremath{\downarrow}      & KLD\ensuremath{\downarrow}        & AKLD\ensuremath{\downarrow}      & KLD\ensuremath{\downarrow}           & AKLD\ensuremath{\downarrow}          \\ \midrule[0.2pt]
NeCA-W$^{\dagger}$        & \underline{0.0513} & \underline{0.1765} & \underline{0.0487} & \underline{0.1819} & \underline{0.0258} & \underline{0.1609} & \underline{0.0603} & \underline{0.1496} & \underline{0.0581} & \underline{0.2205} & \underline{0.0489}                  & \underline{0.1779}        \\
NAFlow$^{\dagger}$        & 0.4482             & 1.6918             & 0.2260             & 1.0135             & 0.3267             & 1.1612             & 0.2200             & 0.8725             & 0.1462             & 0.5456             & 0.2734                  & 1.0569        \\
\textbf{Ours}$^{\dagger}$ & \textbf{0.0350}    & \textbf{0.1567}    & \textbf{0.0312}    & \textbf{0.1553}    & \textbf{0.0222}    & \textbf{0.1392}    & \textbf{0.0291}    & \textbf{0.1229}    & \textbf{0.0204}    & \textbf{0.1326}    & \textbf{0.0276}     & \textbf{0.1413}        \\ \bottomrule[0.5pt]
\end{tabular}
}
\label{table:unpaired_noise_generation_v2}
\end{table*}

\begin{table*}[!t]
\centering
\caption{Quantitative results on denoising generalization performance trained with synthetic noisy-clean pairs generated in an unpaired manner. The best and second-best results are denoted as \textbf{bold} and \underline{underline}, respectively. \textsuperscript{\textdagger} indicates that noise is generated under an unpaired setting.}
\vspace{0.1cm}
\resizebox{0.85\textwidth}{!}{
\begin{tabular}{lcccccccc|cc}
\toprule[0.5pt]
\multirow{2}{*}{Methods} & \multicolumn{2}{c}{PolyU} & \multicolumn{2}{c}{Nam}  & \multicolumn{2}{c}{SIDD Validation} & \multicolumn{2}{c}{SIDD+} & \multicolumn{2}{|c}{Average} \\ \cmidrule(lr){2-3} \cmidrule(lr){4-5} \cmidrule(lr){6-7} \cmidrule(lr){8-9} \cmidrule(lr){10-11}
                         & PSNR$\uparrow$          & SSIM$\uparrow$           & PSNR$\uparrow$          & SSIM$\uparrow$           & PSNR$\uparrow$          & SSIM$\uparrow$           & PSNR$\uparrow$          & SSIM$\uparrow$           & PSNR$\uparrow$          & SSIM$\uparrow$  \\ 
                         \midrule[0.2pt]
NeCA-W$^\dagger$        & \underline{37.39} & \underline{0.9555} & \underline{37.13} & \underline{0.9495} & \underline{32.86} & \underline{0.7490} & \underline{33.92} & \underline{0.8342} & \underline{35.33} & \underline{0.8720} \\
NAFlow$^\dagger$        & 32.40             & 0.9157             & 31.73             & 0.9381             & 30.89             & 0.7776             & 31.60             & 0.7953             & 31.66             & 0.8567             \\
\textbf{Ours}$^\dagger$ & \textbf{38.00}    & \textbf{0.9618}    & \textbf{38.37}    & \textbf{0.9633}    & \textbf{32.93}    & \textbf{0.7803}    & \textbf{34.25}    & \textbf{0.8557}    & \textbf{35.89}    & \textbf{0.8903}    \\
\bottomrule[0.5pt]
\end{tabular}
}
\label{table:unpaired_denoising_result}
\end{table*}

\subsection{Unpaired Noise Generation}
\label{appendix:unpaired_noise_generation}
In~\ttabref{table:noise_gen_per_devices}, we use paired noisy image \text{\footnotesize $\pmb{\mathcal{I}}_\mathrm{Noisy}$} and clean image \text{\footnotesize $\pmb{\mathcal{I}}_\mathrm{Clean}$} 
from test datasets during inference
to compute \text{\footnotesize $\mathrm{\mathbf{n}}_\mathrm{Real}=\pmb{\mathcal{I}}_\mathrm{Noisy}-\pmb{\mathcal{I}}_\mathrm{Clean}$}, following the experimental setup used in previous works~\citep{noiseflow, Flow-sRGB, naflow, neca}. 
To broaden \framework{}’s applicability to cases where only some noisy-clean pairs are available for training alongside extra clean-only images, we also evaluate it in an unpaired setting.
In this scenario, as illustrated in~\ffigref{fig:appx_unpaired_noise_generation}, noise images \text{\footnotesize $\mathrm{\mathbf{n}}_\mathrm{Real}$} are randomly sampled from the training dataset (\eg SIDD) and used to synthesize noisy images \text{\footnotesize $\hat{\pmb{\mathcal{I}}}_\mathrm{Noisy}$} conditioned on new, unpaired clean images \text{\footnotesize $\pmb{\mathcal{I}}_\mathrm{Clean}$} that are randomly sampled from the external dataset (\eg ImageNet~\cite{imagenet}), during the inference.

\subsubsection{Generated Noise Quality Assessment.}
\noindent\textbf{Experiment Settings.} In~\ttabref{table:unpaired_noise_generation_v2}, we evaluate noise quality under the same unpaired setting as described above, using NAFlow, NeCA-W, and~\framework{}. All models are trained on the paired SIDD training dataset, and NeCA-W and NAFlow are evaluated using their official weights as described in~\ssecref{sec:4.2} of the main manuscript.
To evaluate how well noise generation networks model each specific noise domains with SIDD validation, we categorize results by different five smartphone types. Notably, in this table, we use metadata only for categorization; however, in real-world applications, our method can synthesize a dataset following the trained noise distribution without relying on any metadata.

\noindent\textbf{Noise Quality Results.} 
\ttabref{table:unpaired_noise_generation_v2} shows that, while the unpaired setting introduces a slight performance drop compared to the paired setting, \framework{} still achieves comparable or superior noise quality metrics (KLD and AKLD) compared to other baseline methods such as NeCA-W and NAFlow.
Furthermore, \framework{} achieves even superior performance compared to NeCA-W in the paired setting (KLD: 0.0342, AKLD: 0.1436), despite utilizing limited noise information.
This flexibility underscores the adaptability of \framework{}, which synthesizes realistic noisy images without requiring paired clean–noisy images. 

This makes \framework{} also suitable for scenarios where paired datasets are unavailable. Additionally, in the paired setting, \framework{} demonstrates strong performance without the need for explicitly defined metadata such as ISO or camera type. These results further support the practicality of \framework{} across a wide range of use cases, regardless of whether paired or unpaired data are available.

\subsubsection{Denoising Quality Assessment.}
\noindent\textbf{Experiment Settings.}
To further assess the quality of generated noise, we train the DnCNN model only using noisy images generated via an unpaired approach.
To ensure a rigorous assessment, we randomly select noise from 15,000 non-overlapping patches from the SIDD training set and randomly paired these with clean images from the ImageNet~\cite{imagenet} validation set, which contains 50,000 images from various scenes.
Subsequently, 50,000 synthetic noisy-clean image pairs are generated from the ImageNet dataset.

\noindent\textbf{Denoising Results.}
As shown in~\ttabref{table:unpaired_denoising_result}, the DnCNN model trained on our method outperforms SOTA methods, including NeCA-W and NAFlow, across all real-world datasets. 
These results confirm the superior performance of unpaired noise generation with \framework{}. 
Specifically, these denoising results are consistent with the unpaired noise generation results in Table~\ref{table:unpaired_noise_generation_v2}, where NAFlow produces the lowest denoising performance, which can be attributed to its lower-quality unpaired noise generation.
NeCA-W achieves the second-best denoising results, but our method surpasses NeCA-W, with average improvements of 0.56 in PSNR and 0.0183 in SSIM. In summary, compared to recent SOTA methods such as NeCA-W and NAFlow, our approach is capable of generating high-quality noise in both paired and unpaired scenarios, demonstrating its practicality for real-world applications.

\subsection{Qualitative Results of Noise Generation}
\label{appendix:noise_generation_qualitative_evaluation}
\begin{figure*}[t]
    \begin{center}
        \includegraphics[width=0.6\textwidth]{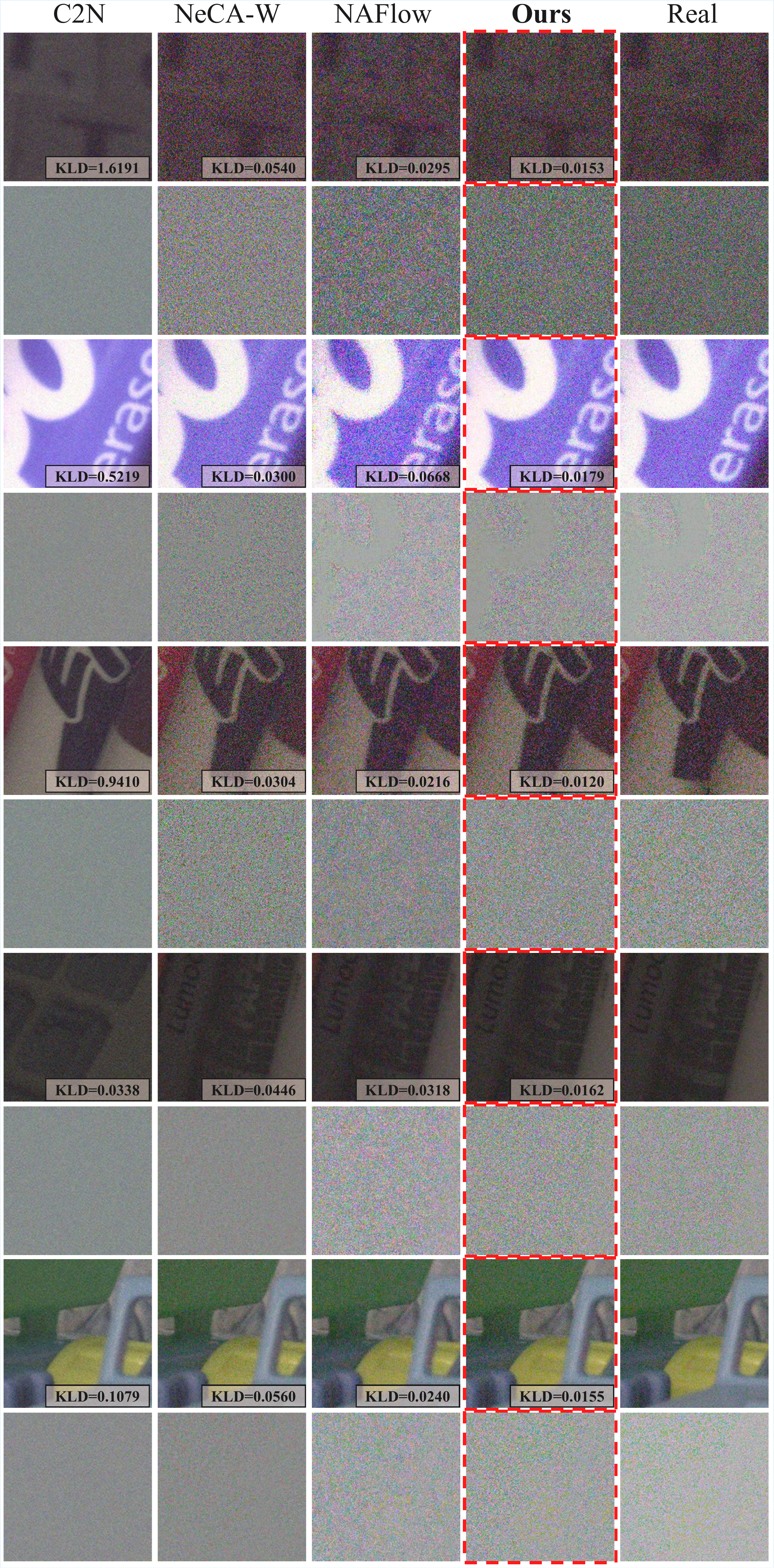}
    \end{center}
    \caption{Visualization of synthetic noisy images on the SIDD validation set. From left to right: C2N, NeCA-W, NAFlow, Ours (\framework{}), and real noisy images.}
    \label{fig:appendix_noise_results}
\end{figure*}

In~\ffigref{fig:appendix_noise_results}, we provide additional visualizations of generated noise, comparing our method (\framework{}) with other approaches: C2N~\citep{c2n}, NeCA-W~\citep{neca}, and NAFlow~\citep{naflow}. 

\subsection{Qualitative Results of Denoising Performance}
\label{appendix:denoising_qualitative_evaluation}
\begin{figure*}[t]
    \begin{center}
        \includegraphics[width=0.7\textwidth]{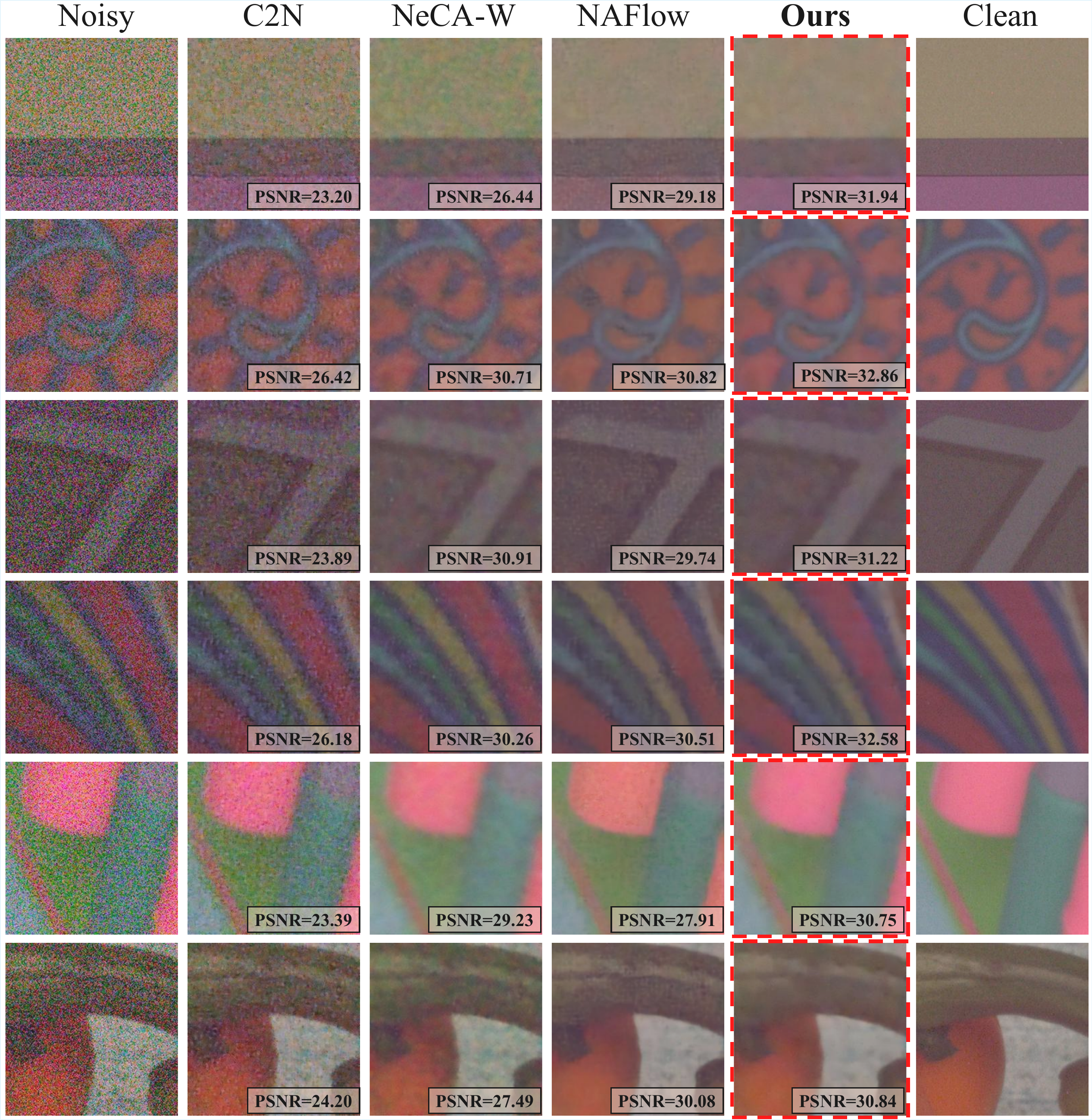}
    \end{center}
    \caption{Visual comparison on denoising results with PSNR$\uparrow$ on SIDD validation set from DnCNN trained on each method.}
    \label{fig:appendix_denoising_results}
\end{figure*}

In~\ffigref{fig:appendix_denoising_results}, we provide additional visualizations of denoising results, where the denoising network is trained on synthetic datasets. We compare our method (\framework{}) with other approaches, including C2N, NeCA-W, and NAFlow.

\clearpage


\end{document}